\documentclass[article,nojss]{jss}

\author{Bryon Aragam\\University of California,\\ Los Angeles \And 
       Jiaying Gu\\University of California,\\ Los Angeles \And
       Qing Zhou\\University of California,\\ Los Angeles}
\title{Learning Large-Scale Bayesian Networks with the \pkg{sparsebn} Package} 

\Plainauthor{Bryon Aragam, Jiaying Gu, Qing Zhou} 
\Plaintitle{Learning Large-Scale Bayesian Networks with the sparsebn Package} 

\Abstract{
Learning graphical models from data is an important problem with wide applications, ranging from genomics to the social sciences. Nowadays datasets often have upwards of thousands---sometimes tens or hundreds of thousands---of variables and far fewer samples. To meet this challenge, we have developed a new \proglang{R} package called \pkg{sparsebn} for learning the structure of large, sparse graphical models with a focus on Bayesian networks. While there are many existing software packages for this task, this package focuses on the unique setting of learning large networks from high-dimensional data, possibly with interventions. As such, the methods provided place a premium on scalability and consistency in a high-dimensional setting. Furthermore, in the presence of interventions, the methods implemented here achieve the goal of learning a causal network from data. Additionally, the \pkg{sparsebn} package is fully compatible with existing software packages for network analysis.
}
\Keywords{Bayesian networks, causal networks, graphical models, machine learning, structural equation modeling, multi-logit regression, experimental data}
\Plainkeywords{Bayesian networks, causal networks, graphical models, machine learning, structural equation modeling, multi-logit regression, experimental data} 

\Address{
  Qing Zhou\\
  UCLA Department of Statistics\\
  Los Angeles, CA 90095, U.S.A.\\
  E-mail: \email{zhou@stat.ucla.edu}\\
  URL: \url{http://www.stat.ucla.edu/~zhou/}
}

\usepackage{amssymb}
\usepackage{amsmath}
\usepackage{amsthm}
\usepackage{enumitem}
\usepackage{algorithm}
\usepackage{booktabs}
\usepackage{subcaption}
\usepackage[multiple]{footmisc}

\usepackage{color}

\DeclareMathOperator*{\pa}{pa}
\DeclareMathOperator*{\diag}{diag}

\newcommand{\rv}{X}
\newcommand{\err}{\varepsilon}
\newcommand{\normalN}{\mathcal{N}}
\newcommand{\trueCov}{\Sigma}
\newcommand{\covest}{\widehat{\trueCov}}
\newcommand{\trueInv}{\Gamma}
\newcommand{\invest}{\widehat{\trueInv}}
\newcommand{\nlevel}{r}
\newcommand{\nll}{\ell}
\newcommand{\R}{\mathbb{R}}
\newcommand{\dagmat}{B}
\newcommand{\varmat}{\Omega}
\newcommand{\dagcomp}{\beta}
\newcommand{\dagest}{\widehat{\dagmat}}
\newcommand{\dagcompest}{\widehat{\dagcomp}}
\newcommand{\varcomp}{\omega}
\newcommand{\varest}{\widehat{\Omega}}
\newcommand{\varcompest}{\widehat{\varcomp}}
\newcommand{\dagspace}{\mathbb{D}}
\newcommand{\reg}{\rho}
\newcommand{\pl}{\reg_{\lambda}}
\newcommand{\samplemat}{\mathbf{X}}
\newcommand{\samplematcol}{\mathbf{x}}
\newcommand{\dummy}{\mathbf{z}}
\renewcommand{\|}{\,|\,}

\begin{document}



\section{Introduction}
\label{sec:intro}

Graphical models are a popular tool in machine learning and statistics, and have been used in a variety of applications including genetics \citep{gao2015,isci2015}, computational biology \citep{jones2012}, oncology \citep{chen2015}, medicine and health care \citep{velikova2014}, logistics \citep{garvey2015}, finance \citep{sanford2012}, and even software testing \citep{dejaeger2013}. The widespread growth of high-dimensional biological data in particular has spurred a renewed interest in the use of graphical models to aid in the discovery of novel biological mechanisms \citep{buhlmann2014bio}. While the past decade has witnessed tremendous developments towards understanding undirected graphical models \citep{meinshausen2006,ravikumar2010,yang2015}, there has been less progress towards understanding directed graphical models---also known as Bayesian networks (BNs) or structural equation models (SEM)---for high-dimensional data with $p\gg n$. A BN is represented by a directed acyclic graph (DAG), whose structure contains a richer and different set of conditional independence relations than an undirected graph. Moreover, DAGs are commonly used in causal inference where the direction of an edge encodes causality. Consequently, there have been continuing efforts in structure learning of directed graphs from data.

Unlike their undirected counterparts, however, the structure learning problem for directed graphical models is complicated by the nonconvexity, nonsmoothness, and nonidentifiability of the underlying statistical problem. These issues have no doubt slowed progress towards fast, scalable algorithms for learning in the presence of thousands---let alone tens of thousands---of variables. Despite progress at the theoretical \citep{geer2013,aragam2016} and computational level \citep{schmidt2007,xiang2013,fu2013}, there is still a lack of user-friendly software for putting these modern tools into the hands of practitioners.

To bridge this gap we have developed \pkg{sparsebn}, a new \proglang{R} \citep{rcore2016} package for structure learning and parameter estimation of large-scale Bayesian networks from high-dimensional data. When experimental data are available, an estimated DAG from this package has a natural causal interpretation. While there are many \proglang{R} packages for learning Bayesian networks (Section~\ref{subsec:packages}), none that we are aware of are specifically tailored to high-dimensional data with experimental interventions. The \pkg{sparsebn} package has been developed from the ground up using recent developments in structure learning \citep{fu2013,aragam2015,gu2018} and statistical optimization \citep{friedman2007,friedman2010,mazumder2011}. In addition to methods for learning Bayesian networks, this package also includes procedures for learning undirected graphs, fitting structural equation models, and is compatible with existing packages in \proglang{R}. All of the code for this package is open-source and available through the Comprehensive \proglang{R} Archive Network (CRAN) at \url{http://CRAN.R-project.org/package=sparsebn}.

To briefly illustrate the use of \pkg{sparsebn}, the code below learns the structure of the pathfinder network \citep{heckerman1992}:
\begin{Sinput}
R> library("sparsebn")
R> data("pathfinder")
R> data <- sparsebnData(pathfinder$data, type = "continuous")
R> dags <- estimate.dag(data)
R> plotDAG(dags)
\end{Sinput}

\begin{figure}[t]
\centering
\includegraphics[width=0.75\textwidth]{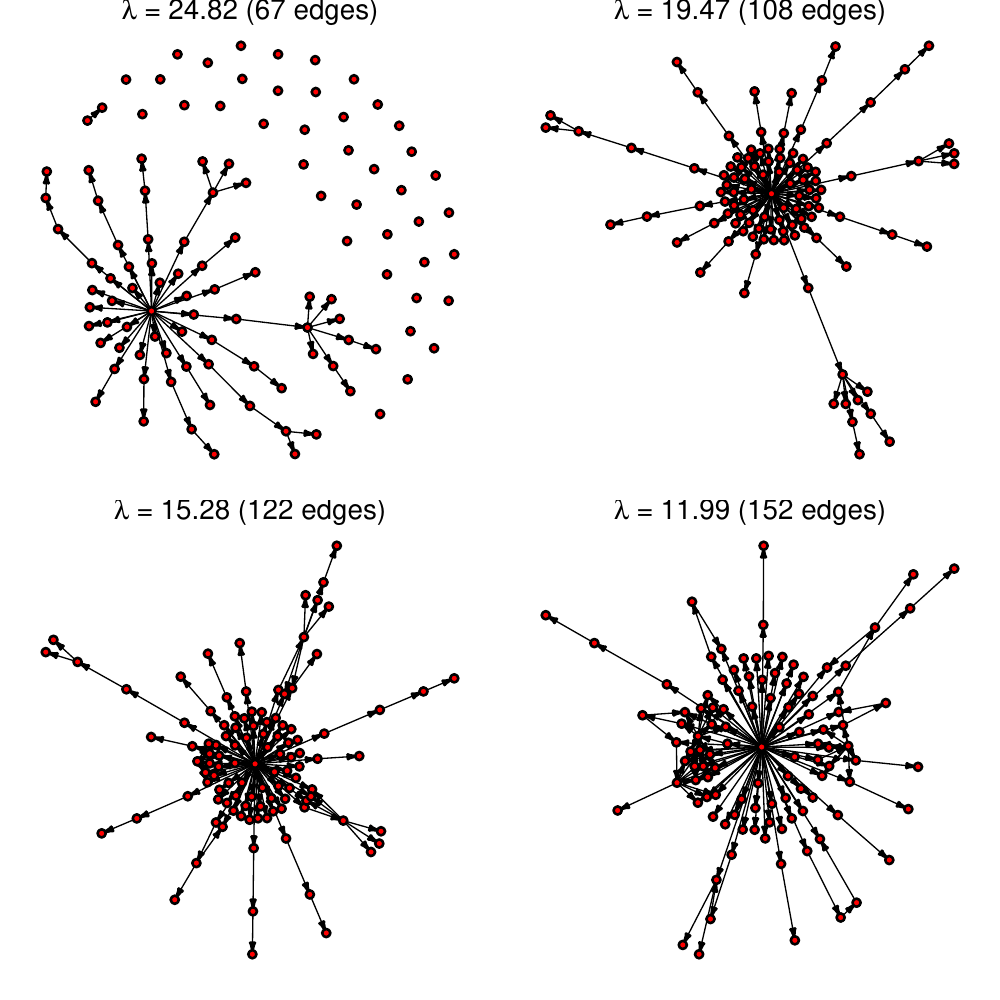}
\caption{Example output from learning the pathfinder network. To save space, only the first four nontrivial estimates are shown here out of the 16 total estimates in the full solution path.}
\label{fig:examplerun}
\end{figure}

\noindent
This code estimates a \emph{solution path} with 16 total estimates (see Section~\ref{subsec:algorithm}) and takes approximately one second to run. The first four estimated networks with an increasing number of edges are shown in Figure~\ref{fig:examplerun}. This example is explored in more detail in Section~\ref{subsec:pathfinder}.

\newpage
\section{Learning Bayesian networks from data}
\label{sec:learningbns}

We begin by reviewing the necessary background and definitions, and then discuss  the existing literature and methods.

\subsection{Background}
\label{subsec:background}

The basic model we work with is a $p$-dimensional random vector $\rv=(\rv_{1},\ldots,\rv_{p})$ with joint distribution $\Prob(\rv_{1},\ldots,\rv_{p})$.
Bayesian networks are directed graphical models whose edges encode conditional independence constraints implied by the joint distribution of $\rv$. For continuous data, we assume that $\rv$ follows a multivariate Gaussian distribution; for discrete data we assume each $\rv_{j}$ is a factor with $\nlevel_{j}$ levels. We do not consider so-called hybrid Bayesian networks that allow for graphs with both continuous and discrete nodes, although this is an interesting future direction for this package.

The general theory of Bayesian networks is quite intricate, and we will make no attempt to cover it in detail here. The interested reader is referred to one of the many excellent textbooks on this subject: \citet{koller2009,spirtes2000,lauritzen1996}. 

Formally, a Bayesian network is defined as a directed acyclic graph $G=(V,E)$ that satisfies the following factorization condition with respect to the joint distribution of $\rv$:
\begin{align*}
\Prob(\rv_{1},\ldots,\rv_{p})
=\prod_{j=1}^{p}\Prob(\rv_{j}\|\pa(\rv_{j}),\theta_{j}).
\end{align*}

\noindent
Here, $\pa(\rv_{j})=\{\rv_{i} : \rv_i\to \rv_j\in E\}$ is the parent set of $\rv_{j}$ and $\theta_{j}$ encodes the parameters that define the conditional probability distribution (CPD) for $X_{j}$.

Traditional methods for learning Bayesian networks start with this definition and develop algorithms from a graph-theoretic perspective \citep{spirtes1991}. This approach comes with restrictive assumptions such as strong faithfulness \citep{uhler2013,zhang2002}, which hinders their use in practice. To motivate our work, we adopt a more general approach to Bayesian networks via structural equation models. In this approach, we start by directly modeling each CPD $\Prob(\rv_{j}\|\pa(\rv_{j}),\theta_{j})$ via a generalized linear model. We will consider two special cases: Gaussian CPDs for continuous data and multi-logit CPDs for discrete data. This approach also fits naturally our framework for learning causal networks discussed in Section~\ref{subsec:causal}.

\subsubsection{Continuous data}

Suppose that for each $j=1,\ldots,p$ there exists $\dagcomp_{j}=(\dagcomp_{1j},\ldots,\dagcomp_{pj})\in\R^{p}$ such that
\begin{align}
\label{eq:gaussian:cpd}
\rv_{j}
= \dagcomp_{j}^{\top}X + \err_{j},
\end{align}

\noindent
where $\dagcomp_{jj}=0$ (to avoid trivialities) and $\err_{j}\sim\normalN(0,\varcomp_{j}^{2})$. By writing $\dagmat=[\dagcomp_{1}\|\cdots\|\dagcomp_{p}]\in\R^{p\times p}$ and $\err=(\err_{1},\ldots,\err_{p})\in\R^{p}$, we can rewrite Equation~\ref{eq:gaussian:cpd} as a matrix equation:
\begin{align}
\label{eq:gaussian:sem}
\rv
= \dagmat^{\top}\rv + \err.
\end{align}

\noindent
The model \eqref{eq:gaussian:sem} is called a \emph{structural equation model} for $\rv$. The matrix $\dagmat$ defines the weighted adjacency matrix of a directed graph which, when acyclic, is also a BN for $\rv$. Note that to estimate $\dagmat$ from data it is not enough to simply regress $\rv_{j}$ onto the rest of variables---this leads to the so-called \emph{neighbourhood regression} estimator of the Gaussian graphical model introduced in \citet{meinshausen2006}. The problem with this approach is that there is no guarantee that the resulting adjacency matrix will be acyclic. In order to produce a BN, we must constrain $\dagmat$ to be acyclic, which couples the parameters of each CPD to one another and induces a nonconvex constraint during the learning phase. For this reason, learning directed graphs is substantially more difficult than learning undirected graphs.

By writing $\varmat=\COV(\err)$, we see that the parameters $(\dagmat,\varmat)$ specify a unique normal distribution $\normalN(0,\trueCov)$ for $\rv$. In fact, a little algebra on Equation~\ref{eq:gaussian:sem} shows that 
\begin{align}
\label{eq:impliedcov}
\trueCov
= (I-\dagmat)^{-\top}\varmat(I-\dagmat)^{-1}.
\end{align}

\noindent
This gives a way to compute $\trueCov$ from $(\dagmat,\varmat)$, and suggests a way to estimate the covariance matrix by setting $\covest= (I-\dagest)^{-\top}\varest(I-\dagest)^{-1}$. By taking $\trueInv=\trueCov^{-1}$,  we obtain an alternative estimator of the Gaussian graphical model, given by 
\begin{align}
\label{eq:impliedprec}
\invest
= (I-\dagest)\varest^{-1}(I-\dagest)^{\top}.
\end{align}

\noindent
This idea was first explored by \citet{rutimann2009} using the PC algorithm (Section~\ref{subsec:previouswork}), and a similar idea using our methods is implemented in \pkg{sparsebn}. In many situations, the DAG representation $(\dagmat,\varmat)$ encodes more conditional independence relations than the inverse covariance matrix $\trueInv$, which is one of the motivations for learning DAGs from observational data.

\subsubsection{Discrete data}

In a discrete Bayesian network, each $X_j$ is a finite random variable with $r_j$ states. Instead of a traditional product multinomial model, \pkg{sparsebn} uses a multi-logit model for discrete data. One of the advantages of this approach is a significant reduction in the number of parameters, and recent work suggests that this model serves as a good approximation to the product multinomial model \citep{gu2018}. Here, we briefly introduce the multi-logit model, and describe how it is used for structure learning.

We use a standard form of the multi-logit model in which each variable $X_j$ is encoded by $d_j = r_j-1$ dummy variables \citep{dobson2008}. More specifically, given a reference category (which may be arbitrary), the $r_{j}$ possible values of $X_{j}$ are encoded by a vector of dummy variables $\mathbf{z}_{j}=(z_{jk},k=1,\ldots,d_j)\in\{0,1\}^{d_{j}}$. Let $I(\,\cdot\,)$ denote the usual indicator function, so that e.g., $I(X_j=k)=1$ if $X_{j}=k$ and $I(X_j=k)=0$ otherwise. If we choose the last category as the reference, then $z_{jk}=I(X_j=k)$ for $k=1,\ldots, d_{j}$ so that the reference category is coded as $\mathbf{z}_j=\mathbf{0}$. 

Under this parametrization, the conditional distributions take the form
\begin{eqnarray}\label{eq:mlogit}
\Prob(X_j=u \| \pa(X_{j})) 
= \dfrac{\exp\big(\beta_{0u j}+\sum_{i\in\pa(X_{j})}\dummy_{i}^{\top}\boldsymbol{\beta}_{iuj}\big)}{\sum_{m=1}^{r_j}\exp\big(\beta_{0u j}+\sum_{i\in\pa(X_{j})}\dummy_{i}^{\top}\boldsymbol{\beta}_{imj}\big)}, \; u=1, \ldots, r_j,
\end{eqnarray}
where $\boldsymbol{\beta}_{iu j} \in \mathbb{R}^{d_i}$ is the coefficient vector for variable $X_i$ to predict the $u$th level of $X_j$ with intercept $\beta_{0u j}$. In this model, if $X_i \notin \text{pa}(X_j)$ then we have equivalently $\boldsymbol{\beta}_{iu j}=\mathbf{0}$ for all $u$. Let 
$\boldsymbol{\beta}_{ij}
=(\boldsymbol{\beta}_{i1j}, \ldots,
\boldsymbol{\beta}_{ir_j j}) \in \mathbb{R}^{d_ir_j}$
be the coefficient vector for edge $i$ to $j$.
 Therefore, by estimating $\boldsymbol{\beta}_{ij}$, we can infer the structure of a network. If $\boldsymbol{\beta}_{ij} = \mathbf{0}$, there does not exist an edge from $X_i$ to $X_j$. For more details on this model, see \citet{gu2018}.

\subsection{Causal DAG learning from interventions}
\label{subsec:causal}

DAGs are a popular model for causal networks, particularly when combined with experimental interventions in addition to observational data \citep{Pearl00}. \citet{cooper1999causal,meganck2006learning,Ellis08} proposed methods to learn causal networks with a mixture of observational and experimental data, and \citet{pe2001inferring,pournara2004reconstruction} inferred gene networks with perturbed expression data.
Each of the methods in \pkg{sparsebn} can take experimental data as input in order to learn the precise causal relationships in a system.

The following simple example illustrates how interventions can be used for learning causal relations: Assume the true causal graph is $\mathcal{G}^*: X_1 \rightarrow X_2$, i.e., $X_{1}$ is a direct cause of $X_{2}$. This is observationally equivalent to the graph $\mathcal{G}: X_1 \leftarrow X_2$---we cannot distinguish between these two graphs using observational data alone. Instead, by manipulating $X_2$ experimentally and fixing its value, we can ``cut off'' the edge from $X_1$ to $X_2$ since the value of $X_{2}$ would no longer be associated with (i.e., is independent of) $X_{1}$. This is an example of an \emph{experimental intervention.} In doing so, the joint distribution factors as $\Prob(X_1)\Prob(X_2)$. Instead, if we manipulated $X_1$, the relation between $X_1$ and $X_2$ would stay the same, and the joint distribution factors as $\Prob(X_1)\Prob(X_2\|X_{1})$. By exploiting these interventions, it is possible to uncover the true causal structure in a physical system.

To see how this can be applied in a statistical setting, we now show how a general form of the density function for experimental data can be derived from the pure observational joint density function. Let $\mathcal{M} \subset\{1,\ldots,p\}$ be the set of variables under intervention, so the joint probability decomposes as
\begin{equation}\label{eq:BNint}
\Prob(X_1,\ldots,X_p)=\prod_{i \notin \mathcal{M}}\Prob(X_i\|\text{pa}(X_i))\prod_{i \in \mathcal{M}}\Prob(X_i\|\bullet),
\end{equation}
where $\Prob(X_i\|\bullet)$ is the marginal distribution of $X_i$ from which experimental samples are drawn. 
Thus, experimental data sets generated from
the true DAG $\mathcal{G}$ can be considered as data sets generated
from a DAG $\mathcal{G'}$, where $\mathcal{G'}$ is obtained by removing all directed edges in $\mathcal{G}$
pointing to the variables under intervention.
Furthermore, we can see that when $\mathcal{M} = \emptyset$,  Equation~\ref{eq:BNint} is the density function for observational data. For more details regarding causal DAG learning, please refer to \citet{Pearl00} and the references therein.

\subsection{Previous work}
\label{subsec:previouswork}

The various algorithms in the literature for structure learning of Bayesian networks fall into three main categories: constraint-based methods, score-based methods and hybrid methods. 

\subsubsection{Constraint-based methods} 
Constraint-based methods rely on repeated conditional independence tests in order to learn the structure of a network. The main idea is to determine which edges cannot exist in a DAG using statistical tests of independence, a procedure which is justified whenever the so-called \emph{faithfulness} assumption holds. These algorithms first use independence tests to learn the skeleton of the network, and then orient $v$-structures along with the rest of the edges. Because of the existence of Markov equivalent DAGs, the direction of some edges may not be decided \citep[for details, see e.g.,][]{koller2009}. The PC algorithm proposed by \citet{spirtes1991} is a well-known example of this kind of method. Another example is the Fast Causal Inference (FCI) algorithm \citep{spirtes2000,colombo2012}, which allows for latent variables in the network. The output of these algorithms is a partially directed graph, which means that there may be some undirected edges in the estimated graph. While the PC algorithm is a powerful method to learn Bayesian networks in low-dimensions with $n$ very large, the performance of the PC algorithm is less competitive in high-dimensions compared to recent score-based methods \citep{aragam2015}.

\subsubsection{Score-based methods}
Score-based methods rely on scoring functions such as the log-likelihood or some other loss function. The goal of these algorithms is to find a DAG that optimizes a given scoring function.
Some popular scoring functions include
several Bayesian Dirichlet metrics \citep{Buntine91, Cooper92,
  Heckerman95}, Bayesian information criterion \citep{Chickering97},
minimum description length \citep{Bouckaert93prob, Suzuki93, Lam94}, and entropy \citep{Herskovits90}. One of the classic score-based methods is the \emph{greedy hill climbing} (HC) algorithm  \citep{russell1995artificial}. This algorithm is fast but tends to predict too many edges in high-dimensional settings. For discrete networks, the K2 algorithm \citep{Cooper92} is another popular method, however, this method requires prior knowledge about the ordering of the network which is often unavailable in applications. There are also Monte Carlo methods \citep{Ellis08, Zhou11,niinimaki2016}, which are quite accurate but also computationally demanding. This limits Monte Carlo methods to smaller networks with only tens of nodes. Each of the learning algorithms implemented in \pkg{sparsebn} is a score-based method, and as such this package represents an attempt to resolve many of the computational and statistical issues cited here with respect to these methods.

\subsubsection{Hybrid methods}
Finally, there are hybrid methods which combine constraint-based and score-based methods. Hybrid methods first prune the search space by using a constraint-based search, and then learn an optimal DAG structure via score-based search \citep{tsamardinos2006max,perrier2008,gamez2011learning}. The max-min hill-climbing (MMHC) algorithm proposed by \cite{tsamardinos2006max} is a powerful method of this kind.

\subsection[Existing packages]{Existing \proglang{R} packages for structure learning}
\label{subsec:packages}

There are several existing \proglang{R} packages for learning and manipulating Bayesian networks. \pkg{bnlearn} is a thorough and actively maintained package that implements a wide variety of classical approaches such as HC and MMHC \citep{scutari2010}. \pkg{pcalg} focuses on the causal interpretation of Bayesian networks, and implements the PC and FCI algorithms along with methods for inferring causal effects \citep{kalisch2012}. Other packages include \pkg{deal} for mixed data \citep{boettcher2003} and \pkg{gRain} \citep{hojsgaard2012} for exact and approximate computations. For structural equation models in particular, \pkg{lavaan} is a modern \proglang{R} package that implements many of the standard fitting procedures for SEM \citep{rosseel2012}. There are also many software packages available on platforms besides \proglang{R},\footnote{For a more complete list, see \url{https://www.cs.ubc.ca/~murphyk/Software/bnsoft.html}} such as \pkg{BEANDisco} \citep{niinimaki2016},\footnote{\url{https://www.cs.helsinki.fi/u/tzniinim/BEANDisco/}} \pkg{AMIDST} \citep{masegosa2017},\footnote{\url{http://www.amidsttoolbox.com/}} and \pkg{GOBNILP} \citep{cussens2017}.\footnote{\url{https://www.cs.york.ac.uk/aig/sw/gobnilp/}}

Motivated by applications to computational biology and machine learning, \pkg{sparsebn} was designed for the following types of applications:
\begin{itemize}
\item Datasets with several thousand variables, which arise in computational biology and machine learning,
\item High-dimensional data with $p\gg n$, which is common in genomics applications with high-throughput datasets, such as gene expression data that have $p\sim 20,000$ and $n\sim 100$,
\item Experimental data with interventions, which is also common in genomics applications.
\end{itemize}

\noindent
Unfortunately, the aforementioned packages either do not scale to handle these types of problems, or cannot process high-dimensional data at all. By contrast, the \pkg{sparsebn} package was specifically designed to fill this gap, with an orientation towards large, high-dimensional, experimental data. In order to achieve this, the methods contained in this package rely on a combination of novel methodology and algorithms in order to scale to larger and larger datasets. Of course, these methods also gracefully degrade to handle simpler settings with observational and/or low-dimensional data. Moreover, whenever possible, cross-compatiblity with the above mentioned packages has been provided (Section~\ref{subsec:compat}).

\section{Learning with sparse regularization}
\label{sec:methods}

To learn a Bayesian network from data, we use a score-based approach based on regularized maximum likelihood estimation that allows for the incorporation of experimental data. In this section we discuss these details as well as the block coordinate descent algorithm used to approximate the resulting optimization problem.

\subsection{Regularized maximum likelihood}
\label{subsec:pmle}

Suppose $\samplemat\in\R^{n\times p}$ is a matrix of observations, and let $\nll$ denote the negative log-likelihood and $\pl$ be some regularizer. For example, $\pl$ may be the $\ell_{1}$ penalty \citep{tibshirani1996}, the group norm penalty \citep{yuan2006}, or a nonconvex penalty such as the smoothly clipped absolute deviation (SCAD) \citep{fan2001} or minimax concave penalty (MCP) \citep{zhang2010}. Furthermore, we assume that $\samplemat$ does not contain any missing values. If the data contains missing values, these should be imputed first.

We consider the following program:
\begin{align}
\label{eq:mainproblem}
\min_{\dagmat\in\dagspace} \nll(\dagmat;\,\samplemat) + \pl(\dagmat),
\end{align}

\noindent
where $\dagspace\subset\R^{p\times p}$ is the set of weighted adjacency matrices that represent directed graphs without cycles. The resulting problem \eqref{eq:mainproblem} is a nonconvex program, where the nonconvexity arises from (potentially) all three terms: The constraint $\dagspace$, the loss function $\nll$, and the regularizer $\pl$.

For continuous data, we use a Gaussian likelihood derived from Equation~\ref{eq:gaussian:sem} combined with $\ell_{1}$ or MCP regularization, and for discrete data we use a multi-logit model as in Equation~\ref{eq:mlogit} combined with a group lasso penalty. When some data are generated under experimental intervention, we can derive the form of $\nll(\dagmat;\,\samplemat)$ using the strategy discussed in Section~\ref{subsec:causal}: When $X_i$ is under intervention, its marginal distribution is known and hence can be considered constant in Equation~\ref{eq:BNint}. Recalling that $\mathcal{M} \subset\{1,\ldots,p\}$ is the set of variables under intervention, it follows that
\begin{align*}
\Prob(X_1,\ldots,X_p) \propto \prod_{i \notin \mathcal{M}}\Prob(X_i\|\text{pa}(X_i)).
\end{align*}

\noindent
Let $\mathcal{I}_j$ be the set of row indices of the data matrix $\mathbf{X}$ where node $X_j$ is under intervention, and $\mathcal{O}_j = \{1, \ldots, n\} - \mathcal{I}_j$ be the collection of observations for which $X_j$ is not under intervention. Then, according to Equation~\ref{eq:BNint}, the negative log-likelihood factorizes as 
\begin{align}\label{eq:intervL}
\nll(\dagmat;\,\samplemat) = -\sum_{j=1}^p\sum_{h\in \mathcal{O}_j}\log f_{\beta_{j}}(\samplematcol_{hj}\|\text{pa}(\samplematcol_{hj})),
\end{align}

\noindent
where $\dagmat = [\,\beta_{1}\|\cdots\|\beta_{p}\,]$, $f_{\beta_{j}}$ is the conditional density for the $j$th node, and $\samplematcol_{hj}$ is the value of node $X_j$ at the $h$th data point. Note that multiple nodes may be intervened for a particular data point. By incorporating experimental interventions in this way, we are able to orient the edges in a causal Bayesian network and thus distinguish between equivalent DAGs. A similar strategy from a Bayesian perspective was first adopted in \citet{cooper1999causal}. Evidently, one advantage of this framework is its universal applicability to different data types and various likelihood models. 

Since the program \eqref{eq:mainproblem} is nonconvex it is generally regarded as infeasible to find an exact global minimizer of this problem. Indeed, score-based structure learning is known to be NP-hard \citep{chickering2004}. Instead, we seek local minimizers of \eqref{eq:mainproblem} through an optimization scheme based on block coordinate descent. The method is based on the following observation: The difficulty in solving \eqref{eq:mainproblem} lies in enforcing the constraint $\dagmat\in\dagspace$, which is a highly nonconvex and singular parameter constraint. Instead, if we consider \eqref{eq:mainproblem} \emph{one} edge at a time, the problem simplifies considerably. This is the heuristic exploited by many score-based methods (most notably greedy hill climbing). Unlike conventional methods, however, \citet{fu2013} propose a \emph{block-cyclic} strategy and show that it outperforms existing approaches based on greedy updates. This general observation has been exploited to construct the family of fast algorithms implemented in \pkg{sparsebn}.

\subsection{Algorithm details}
\label{subsec:algorithm}

Recalling that $\dagmat=(\dagcomp_{ij})$, the high-level idea behind the method is the following:

    \begin{itemize}[leftmargin=5mm, itemsep=0mm]
    \item[1.] Repeat outer loop until stopping criterion met:
    \item[2.] \emph{Outer loop.} For each pair $(j,k)$, $j\ne k$:
    \begin{itemize}[leftmargin=3mm, itemsep=0mm]
    \item[(a)] Minimize \eqref{eq:mainproblem} with respect to $(\beta_{kj},\beta_{jk})$, holding all other parameters fixed; 
    \item[(b)] If the edge $k\to j$ (resp. $j\to k$) induces a cycle in the graph, set ${\beta}_{kj}\leftarrow 0$
    (resp. ${\beta}_{jk}\leftarrow 0$) and then update $\beta_{jk}$ (resp. $\beta_{kj}$);
    \item[(c)] Repeat inner loop until convergence:
    \end{itemize}
    \begin{itemize}[itemsep=0mm]
    \item[3.] \emph{Inner loop.}
    Fix the edge set $E$ from the outer loop and minimize \eqref{eq:mainproblem} by cycling through the edge weights $\beta_{kj}$ for $(k,j)\in E$.
    \end{itemize}
    \end{itemize}

Since the program \eqref{eq:mainproblem} depends on the unknown regularization parameter $\lambda$, this value must be supplied in advance to the algorithm. In practice, we wish to solve \eqref{eq:mainproblem} for several values of the regularization parameter, so instead of returning a single DAG estimate, the output of this meta-algorithm is a \emph{solution path} \citep[also called a \emph{regularization path},][]{friedman2010}: A sequence of estimates $\{\dagest(\lambda_{\max}),\dagest(\lambda_{1}),\ldots,\dagest(\lambda_{\min})\}$ for a pre-determined grid of values $\lambda_{\max}>\lambda_{1}>\cdots>\lambda_{\min}$. This is standard practice in the literature on coordinate descent \citep{friedman2007,wu2008}, and is similar to the well-known graphical lasso for undirected graphs \citep{friedman2008}.

As $\lambda$ decreases, there is less regularization, hence the resulting estimates $\dagest(\lambda_{m})$ become more dense (i.e., contain more edges). Since our focus is on sparse graphs, in practice we use this fact to terminate the algorithm early if the solution path becomes too dense, i.e., if the number of edges in $\dagest(\lambda_{m})$ exceeds some user-defined threshold. The default values used in \pkg{sparsebn} are $10p$ edges for continuous data and $3p$ edges for discrete data. The smaller threshold for discrete data is due to the higher computational complexity of the underlying algorithm as compared with the continuous case.

The detailed implementation of the algorithms uses several tricks from the literature on coordinate descent in order to speed up the algorithm:
\begin{itemize}
\item \emph{Warm starts.} Given the previous estimate $\dagest(\lambda_{m-1})$ in the solution path, we use $\dagest(\lambda_{m-1})$ as the initial guess for the next iterate $\dagest(\lambda_{m})$. Furthermore, it is always possible to choose $\lambda_{\max}$ so that $\dagest(\lambda_{\max})=\mathbf{0}$ (i.e., the zero matrix), which is the default implementation used by \pkg{sparsebn}.
\item \emph{Active set iterations.} In the inner loop above, the algorithm only updates the nonzero parameters by solving at most $p$ penalized regression or multi-logit regression problems. These subproblems are computationally tractable and can be solved efficiently, which yields significant performance improvements for large graphs.
\item \emph{Block updates.} Instead of updating each $\beta_{jk}$ one at a time, the algorithms update each parameter as a block $\{\beta_{jk},\beta_{kj}\}$. This is justified by the acyclicity assumption: If $\beta_{jk}\ne 0$ and $\beta_{kj}\ne 0$, then the acyclic constraint is violated, and this fact is exploited to update both parameters simultaneously.
\item \emph{Sparse data structures.} Internally, everything is stored using sparse data structures for representing directed acyclic graphs. This saves memory and speeds up the computation of each update.
\end{itemize}

\noindent
Compared with existing packages for structure learning, the main novelties of the present methods are 1) The use of $\ell_{1}$ and MCP regularization, 2) Block-cyclic (as opposed to greedy, one-at-a-time) updates, and 3) The use of warm starts in computing the solution path. Further details can be found in \citet{fu2013} and \citet{aragam2015}.

\subsection{Parameter estimation}
\label{subsec:paraminference}

After learning the structure of a Bayesian network, often it is of interest to then estimate the parameters of the local conditional distributions associated with the learned structure. For causal DAGs, these parameters determine the causal effect sizes between the parents and their children.

For Gaussian data, this is straightforward by regressing each node onto its parents, using (unpenalized) least squares regression. Note that this requires that the maximum number of parents is at most $n$, which is another motivation for leveraging sparsity in our algorithms. This produces a \emph{weighted} adjacency matrix $\dagest=(\dagcompest_{ij})$ (or more specifically, a solution path of adjacency matrices). Given these weights, we can estimate the conditional variance of each node given its parents by:
\begin{align*}
\varcompest_{j}^{2}
:= \VAR(\samplematcol_{j} - \samplemat\dagcompest_{j}).
\end{align*}

\noindent
This yields a variance matrix $\varest=\diag(\varcompest_{1}^{2},\ldots,\varcompest_{p}^{2})$, and together $(\dagest,\varest)$ can be used to estimate the covariance matrix $\trueCov$ (see Section~\ref{subsec:background}).

For discrete data, we regress each node onto its parents set using multi-logit regression via the \proglang{R} package \textbf{nnet} \citep{venables2002}.  
Note that for discrete data, instead of a coefficient matrix we get a 4-way array $\widehat{B} = (\widehat{\boldsymbol{\beta}}_{i j})$, where $\widehat{\boldsymbol{\beta}}_{ij}$ is a matrix and the $(u,k)$ entry of this matrix is the influence that level $k$ of $X_i$ has on level $u$ of $X_j$.

\section[The sparsebn package]{The \pkg{sparsebn} package}
\label{sec:sparsebn}

Based on the framework described in Section~\ref{sec:methods}, \pkg{sparsebn} implements four structure learning algorithms that are each tailored to a specific type of data:
\begin{itemize}
\item Experimental, Gaussian data \citep{zhang2016},
\item Experimental, discrete data \citep{gu2018},
\item Observational, Gaussian data \citep{aragam2015},
\item Observational, discrete data \citep{gu2018}.
\end{itemize}

\noindent
By combining these approaches, \pkg{sparsebn} automatically handles datasets with mixed observational and experimental data. Each algorithm is implemented in \proglang{C++} using \pkg{Rcpp} \citep{eddelbuettel2011,eddelbuettel2013}. In addition to the main algorithms, the package implements high-level methods for fitting, plotting, and manipulating graphical models. 

Furthermore, \pkg{sparsebn} is actually a family of \proglang{R} packages, designed to be cross-compatible with minimal external dependencies. To date, \pkg{sparsebn} imports the following packages:
\begin{itemize}
\item \pkg{ccdrAlgorithm}, based on \citet{aragam2015} and \citet{fu2013},
\item \pkg{discretecdAlgorithm}, based on \citet{gu2018},
\item \pkg{sparsebnUtils}, for housing various common utilities and classes.
\end{itemize}

\noindent
The idea is that the codebase for each algorithm is housed inside its own package, allowing for rapid development and convenient extensibility. This allows us to add new algorithms and features rapidly without significant dependency or compatibility constraints. 

\subsection{Speed and scalability improvements}

\begin{figure}[t]
\centering
\includegraphics[width=0.9\textwidth]{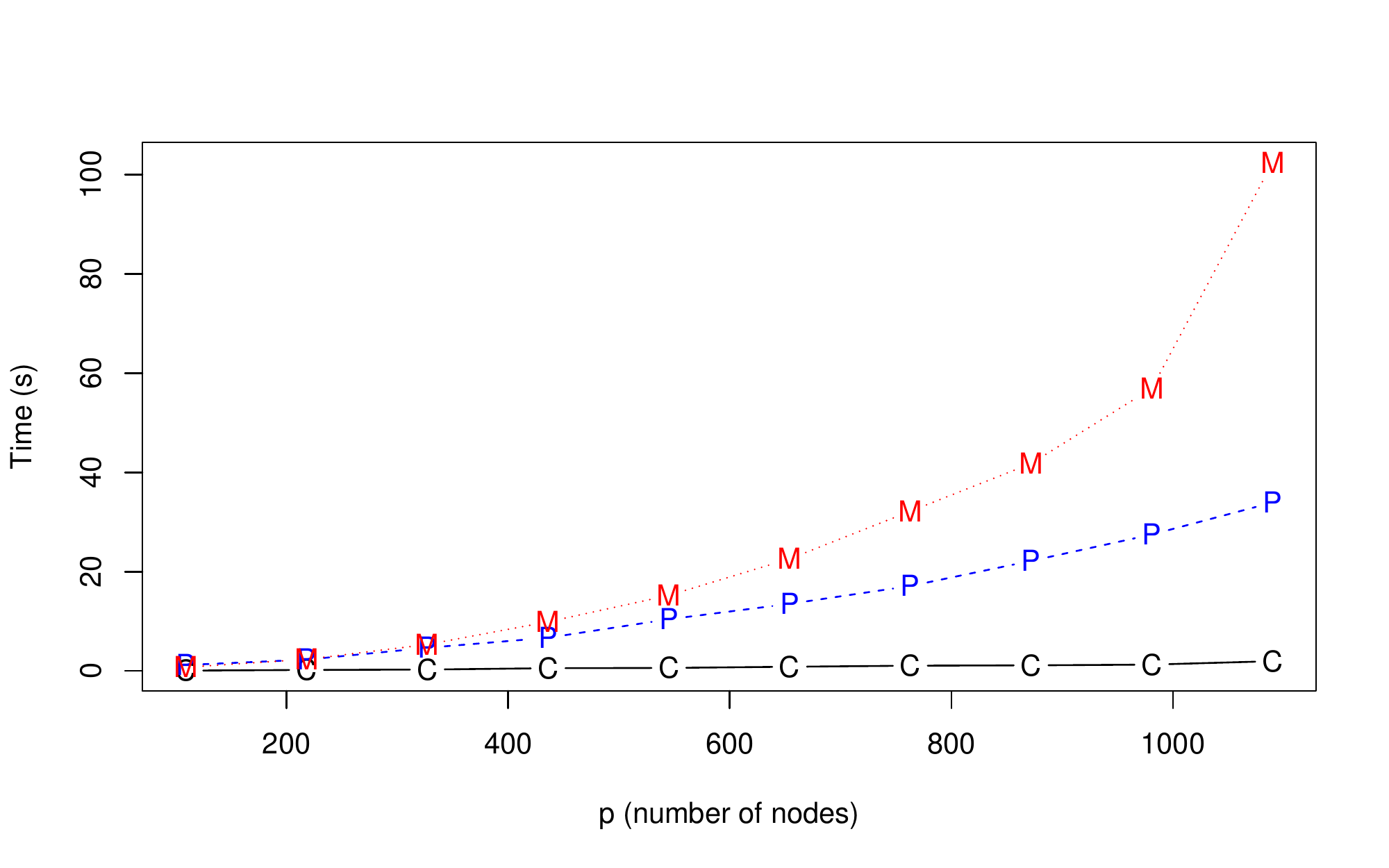}
\caption{Timing comparison (in seconds). Each point represents the total runtime to execute an algorithm on simulated Gaussian data as a function of the number of nodes $p=109k$ ($k=1,\ldots,10$). The DAGs were constructed by tiling $k$ independent copies of the pathfinder network ($p=109$), and $n=50$ samples were randomly generated for each dataset.
(solid black line) C = CCDr algorithm implemented in \pkg{sparsebn}, (dashed blue line) P = PC algorithm implemented in \pkg{pcalg}, (dotted red line) M = MMHC algorithm implemented in \pkg{bnlearn}.}
\label{fig:scalability}
\end{figure}

\pkg{sparsebn} is designed to handle large, high-dimensional datasets with $p$ potentially in the thousands. To illustrate this, Figure~\ref{fig:scalability} provides a comparison of our methods with the PC algorithm from the \pkg{pcalg} package and the MMHC algorithm from the \pkg{bnlearn} package. Due to the higher computational workload for discrete algorithms in general, the results here are for Gaussian data only. Furthermore, although these numbers are intended to be illustrative, the interested reader may find more extensive simulations corroborating these results for both types of data in \citet{aragam2015,gu2018}.

In order to provide a direct comparison, the times reported in Figure~\ref{fig:scalability} reflect the total time to learn graphs of the same complexity (number of edges), even though \pkg{sparsebn} is capable of computing further into the solution path if desired. Note also that both the PC and MMHC algorithms output only a single graph, whereas our method outputs a solution path with multiple graph estimates. The results illustrate how \pkg{sparsebn} provides a more favourable scaling when $p$ is large: For the largest graph with 1090 nodes, \pkg{sparsebn} is 17x faster compared to the PC algorithm in \pkg{pcalg} and 52x faster compared to the MMHC algorithm in \pkg{bnlearn}. These computational improvements are made possible by the use of an efficient block coordinate descent algorithm that leverages warm starts and active set updates; for more details see Section~\ref{subsec:algorithm}.

\subsection{Experimental interventions}

\begin{figure}[t]
\centering
\begin{minipage}[b]{0.45\linewidth}
\centering
\includegraphics[width=1\linewidth]{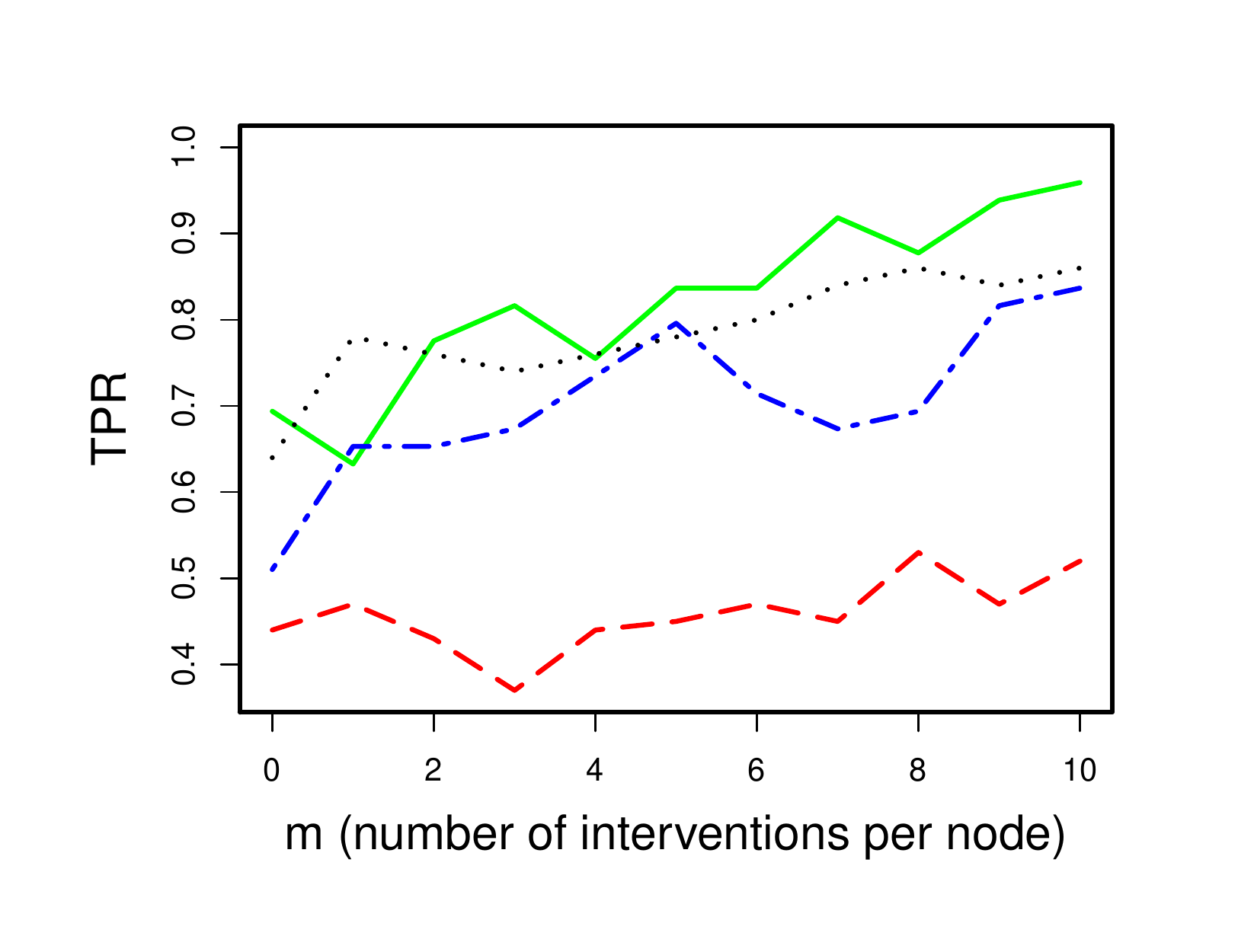} \\
(A)
\end{minipage}
\begin{minipage}[b]{0.45\linewidth}
\centering
\includegraphics[width=1.1\linewidth]{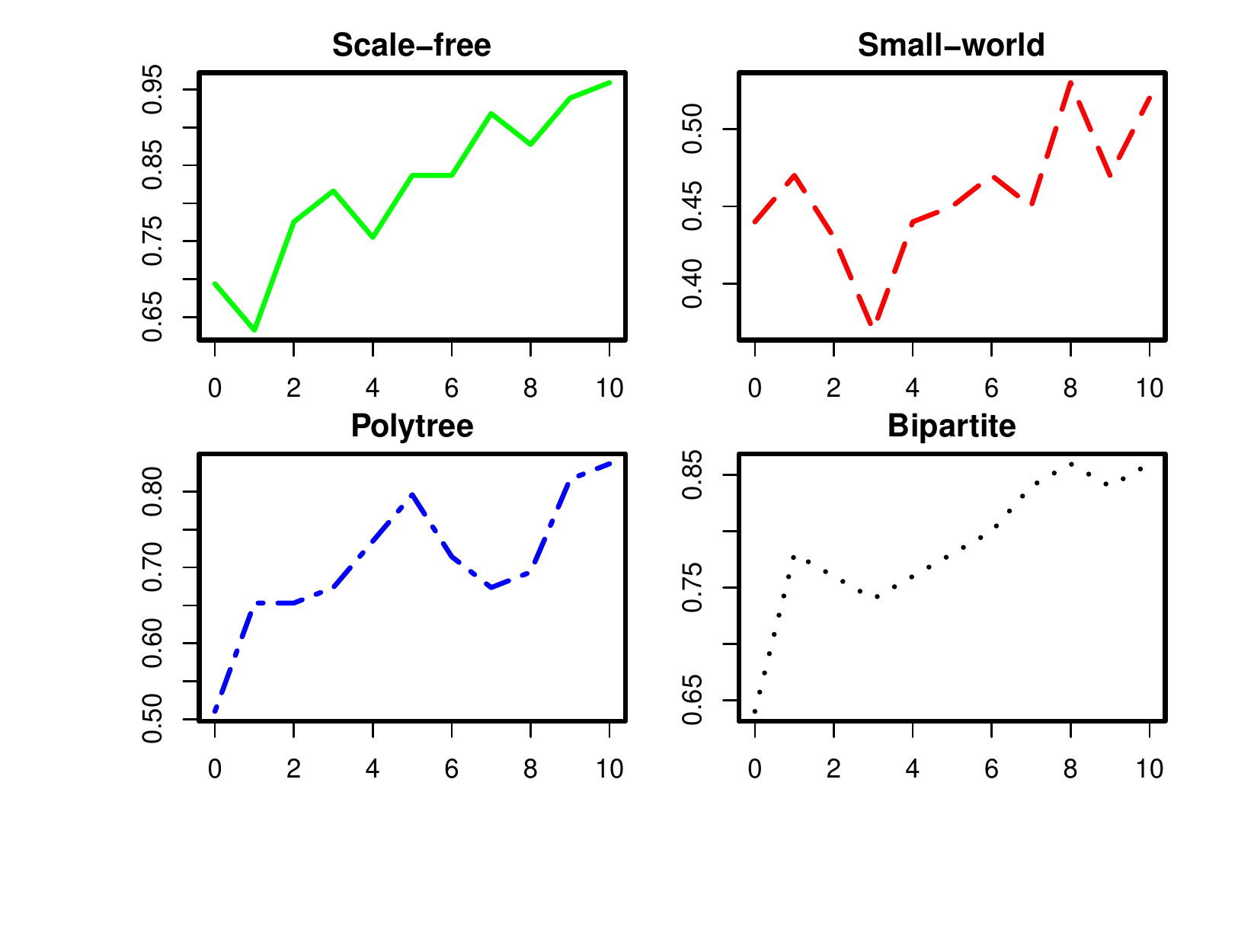} \\
(B)
\end{minipage}
\caption{The effect of interventions in learning discrete DAGs. For every node, $m$ interventions are added to otherwise observational data with $n = 500$ and $p = 50$. (A) Scale-free network with 49 edges (solid green line), small-world network with 100 edges (dashed red line), polytree with 49 edges (dashed blue line), and bipartite graph with 50 edges (dashed black line); (B) Plot for each network individually.
}
\label{fig:intervene-all}
\end{figure}

\begin{figure}[t]
\centering
\begin{minipage}[b]{0.45\linewidth} 
\centering
\includegraphics[width=1\linewidth]{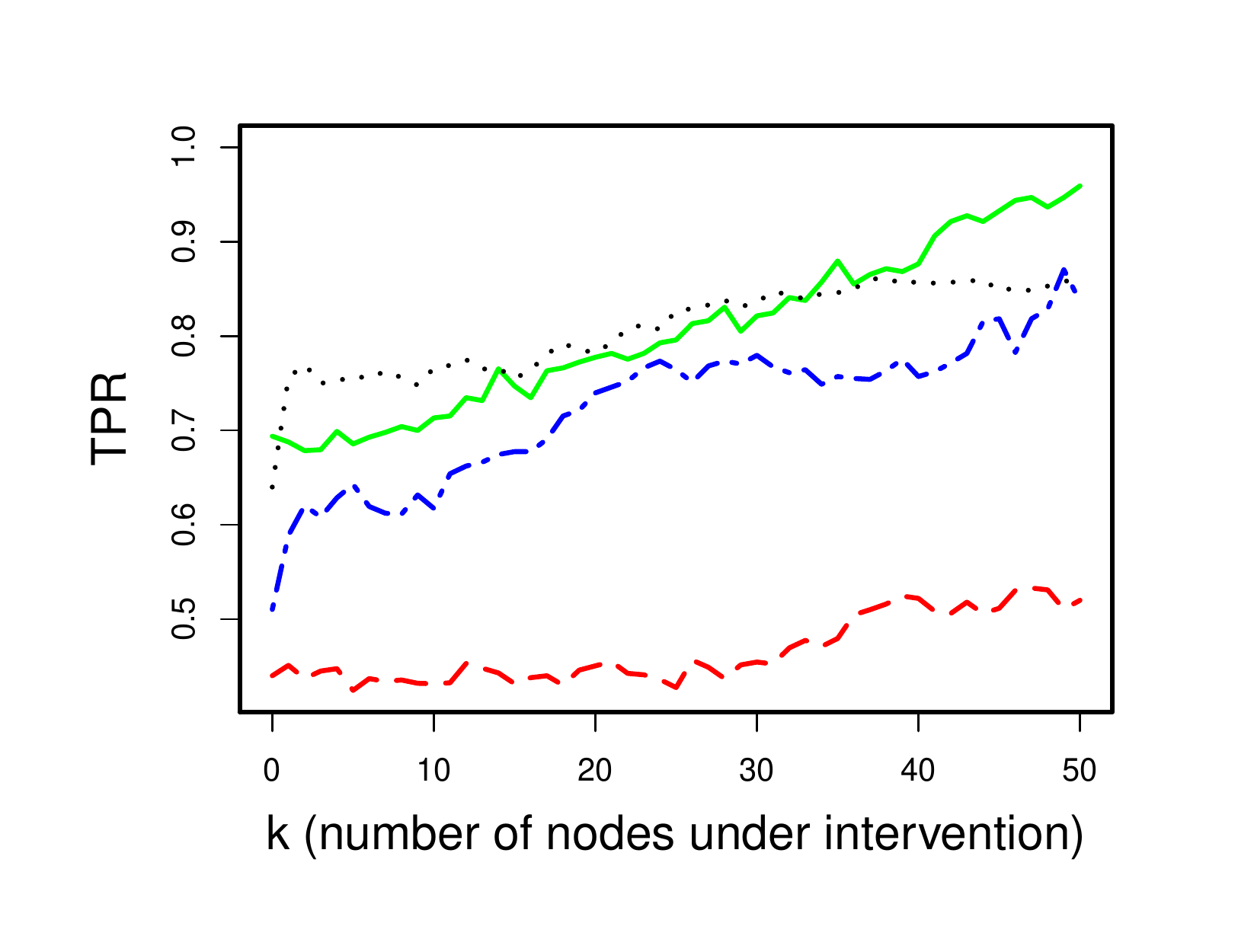} \\
(A)
\end{minipage}
\begin{minipage}[b]{0.45\linewidth}
\centering
\includegraphics[width=1.1\linewidth]{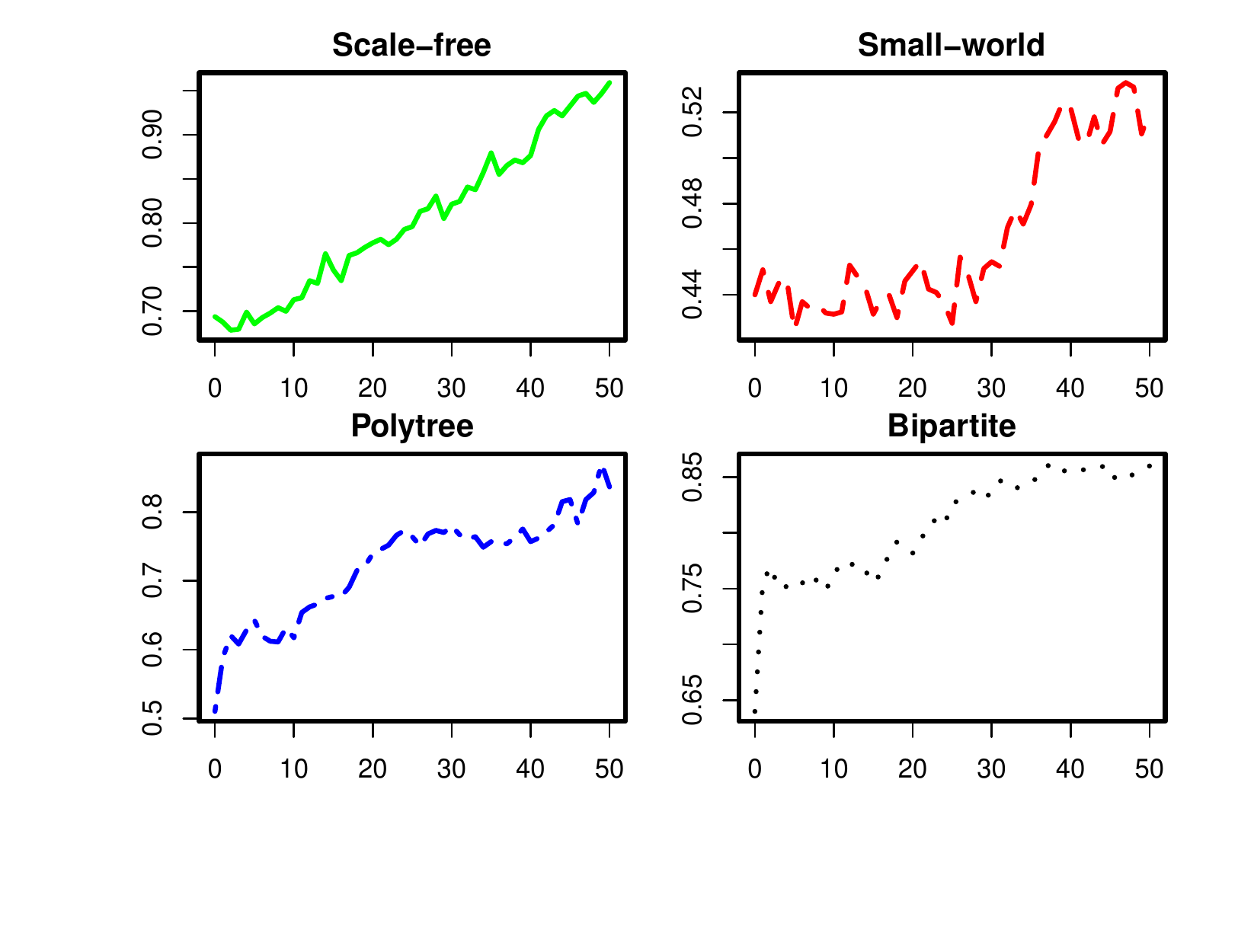} \\
(B)
\end{minipage}
\caption{The effect of interventions in learning discrete DAGs. For each $k=0,\ldots,50$, $m=10$ interventions for $k$ randomly selected nodes are added to otherwise observational data with $n = 500$ and $p = 50$. Results are averaged over 20 random permutations of the order in which each node is intervened on as $k$ is increased. (A) Scale-free network with 49 edges (solid green line), small-world network with 100 edges (dashed red line), polytree with 49 edges (dashed blue line), and bipartite graph with 50 edges (dashed black line); (B) Plot for each network individually.}
\label{fig:intervene-sub}
\end{figure}

In addition to scalability, another feature that distinguishes \pkg{sparsebn} is its native support for mixed observational and experimental data. As discussed in Section~\ref{subsec:causal}, experimental interventions allow observationally equivalent DAGs to be distinguished, thereby uncovering the structure of the true causal DAG. To illustrate this, we ran two simulation experiments, reported in Figures~\ref{fig:intervene-all} and~\ref{fig:intervene-sub}. To keep things simple, we focus on discrete data.\footnote{Code to run these experiments on continuous data can be found at \url{https://github.com/itsrainingdata/sparsebn-reproduce}.} As in the previous subsection, a more thorough evaluation of the effect of interventions can be found in \citet{gu2018,zhang2016,fu2013}.

Figure~\ref{fig:intervene-all} illustrates how the accuracy of reconstruction improves as interventions are added to each node in the network. We can see that as we process more interventions per node (denoted by $m$) with $n$ held fixed, the true positive rate (TPR), increases. Further analysis of these results indicates furthermore that the number of reversed edges decreases along with the number of false positives, and as a result the overall structural Hamming distance (SHD)---the total number of edge additions, deletions, and reversals needed to convert one directed graph into another---decreases. 

Figure~\ref{fig:intervene-sub} considers the more practical scenario in which only $k$ ($k<p$) nodes in the network are under intervention, and over time more interventions on more nodes are able to be collected. This is common, for example, when reconstructing large networks in biological applications. Again, we see that overall the true positive rate increases as more nodes are under intervention, and further analysis shows that the SHD decreases as well.

Since $n$ is held fixed in each simulation, the improvements observed here cannot be attributed to an increase in sample size, illustrating how \pkg{sparsebn} is able to improve estimation of the true causal graph under experimental interventions. In particular, reducing reversed edges that are observationally equivalent shows in an obvious way the utility of interventions for causal inference.

\subsection{Functions}
\label{subsec:functions}

The main purpose of the \pkg{sparsebn} package is estimation of graphical models, which is accomplished through the methods prefaced with ``\code{estimate.}''. In this section, we present an overview of the main estimation methods. In addition, \pkg{sparsebn} includes methods for generating random graphs, simulating random data, visualization, and conversion between various graph classes, which are documented extensively in the package manual.

The main function is \code{estimate.dag}, which can be called as follows:
\begin{Sinput}
R> estimate.dag(data, lambdas = NULL, lambdas.length = 20, whitelist = NULL,
    blacklist = NULL, error.tol = 1e-04, max.iters = NULL,
    edge.threshold = NULL, concavity = 2, weight.scale = 1, convLb = 0.01,
    upperbound = 100, adaptive = FALSE, verbose = FALSE)
\end{Sinput}

\noindent
The main arguments are \code{data}, \code{lambdas}, and \code{lambdas.length}. By default, the \code{lambdas} argument is \code{NULL} and a standard sequence of $L=20$ regularization parameters is generated. If desired, the user can pre-compute a vector of regularization parameters to be used instead, in which case this vector should be passed through the \code{lambdas} argument. If there is prior knowledge of (directed) edges that are known to be present in the network, these can be specified via the \code{whitelist} argument. Similarly, if there is prior knowledge of (directed) edges that are known to be absent from the network, these can be specified via the \code{blacklist} argument. The rest of the arguments control the convergence of the internal algorithms, and are  intended for advanced users. This method returns a \code{sparsebnPath} object (Section~\ref{subsec:ds}), which stores the solution path described in Section~\ref{subsec:algorithm}. 

It is important to bear in mind that the objects returned by \code{estimate.dag} are graphs, and in particular they do not include estimates of model parameters such as edge weights or conditional variances. To obtain these parameters, \pkg{sparsebn} includes the \code{estimate.parameters} method, which can be called as follows:
\begin{Sinput}
R> estimate.parameters(fit, data, ...)
\end{Sinput}

\noindent
where \code{fit} is the output of \code{estimate.dag} and \code{data} is the data to be used for parameter estimation.

In addition to estimating DAGs, \pkg{sparsebn} can estimate the precision and/or covariance matrix for multivariate Gaussian data. The nonzero entries in the precision matrix in particular yield the so-called \emph{Gaussian graphical model}, which is an undirected graphical model for multivariate Gaussian data. This can be done via the \code{estimate.precision} and \code{estimate.covariance} methods:
\begin{Sinput}
R> estimate.covariance(data, ...)
R> estimate.precision(data, ...)
\end{Sinput}

\noindent
Internally, these methods call \code{estimate.dag} and use Equation~\ref{eq:impliedprec} to compute the estimated precision (or covariance) matrix. The \code{...} argument here allows the user to specify any of the optional arguments from \code{estimate.dag}.

\subsection{Data structures}
\label{subsec:ds}

\pkg{sparsebn} uses three different S3 classes in order to represent data (\code{sparsebnData}), graphs (\code{sparsebnFit}), and solution paths (\code{sparsebnPath}). For each of these classes, the usual generics are defined such as \code{print}, \code{summary}, and \code{plot}.

The \code{sparsebnData} class is used to represent both continuous and discrete data with experimental interventions. Observational data corresponds to the degenerate case where $\samplemat$ does not contain any interventions, and is treated as such by the \pkg{sparsebn} package. The slots are:
\begin{itemize}
\item \code{data}: This is the original data as a data frame with $n$ observations and $p$ variables.
\item \code{type}: Either \code{"continuous"} or \code{"discrete"}.
\item \code{levels}: A list of levels for each variable. This is a list of length $p$ whose $j$th component is a vector containing the levels of the $j$th variable.
\item \code{ivn}: The list of interventions for each observations. This is a list of length $n$ whose $i$th component is a vector of node names (or indices) that are under intervention for the $i$th observation.
\end{itemize}

The \code{sparsebnPath} class represents a solution path, which is the output of the main function \code{estimate.dag}. Internally, this is a \code{list} of \code{sparsebnFit} objects whose $j$th component corresponds to the $j$th value of $\lambda$ in the solution path, $\lambda_{\max}>\lambda_{1}>\cdots>\lambda_{\min}$. Since this class is essentially a wrapper for this list, it has no named slots.

The \code{sparsebnFit} class represents an individual graph estimate from a DAG learning algorithm. The graph itself is stored as an \code{edgeList} object in the \code{edges} slot, which is an internal implementation of a child-parent edge list. Alternatively, this graph can also be stored in a variety of other formats including \code{graphNEL} (from the \pkg{graph} package), \code{igraph} (from the \pkg{igraph} package), and \code{network} (from the \pkg{network} package) by using the \code{setGraphPackage} method (Section~\ref{subsec:compat}). The slots are:

\begin{itemize}
\item \code{edges}: A directed graph corresponding to the estimated network, stored internally as an \code{edgeList} by default.
\item \code{nodes}: A vector of node names for the graph.
\item \code{lambda}: The value of $\lambda$ used to estimate the network.
\item \code{nedge}: The total number of edges in the graph.
\item \code{pp, nn, time}: The number of nodes, number of samples, and clock time to estimate this network (these are mainly used internally by the package).
\end{itemize}

\subsection{Compatibility}
\label{subsec:compat}

Unfortunately, there is no consistent standard in \proglang{R} for storing and representing graphs. As a result, different domains have adopted different \proglang{R} packages as a \emph{de facto} standard for graph and network representation. For example, in biology the \pkg{graph} package \citep{gentleman2016} seems to be the most popular, whereas in social science and demography the \pkg{network} package \citep{butts2008} is more popular. In other domains, the \pkg{igraph} package \citep{csardi2006} is popular, which has libraries in \proglang{R}, \proglang{Python}, and \proglang{C}. 

For this reason, \pkg{sparsebn} does not provide its own mechanism for manipulating graphs, and instead provides cross-compatibility with each of these three packages. By default, all methods output graphs stored as an \code{edgeList} object, which is an internal class with little built-in functionality outside of being a storage mechanism for graph data. In order to make use of the extensive capabilities of the different graph packages in \proglang{R}, we have included the \code{setGraphPackage} method to allow users to set a global preference for which graph package to use. Once this preference is set, the full feature set of the selected package (e.g., plotting, manipulation, network statistics, etc.) becomes available to the user. We emphasize that the purpose of \pkg{sparsebn} is not to provide a new library for graph representation and visualization, but instead to provide algorithms for learning their structure. The manipulation of graphs is appropriately left to libraries designed explicitly for that purpose.

Furthermore, to allow cross-compatibility with existing packages for structure learning, we have provided methods to convert the output of \pkg{sparsebn} methods to \pkg{bnlearn}-compatible objects. Compatibility with the \pkg{pcalg} package is possible via the aformentioned \pkg{graph} package, which is the default representation used by \pkg{pcalg}.

The \code{setGraphPackage} method sets a \emph{global} preference for the underlying class used to store graphs by the package. That is, all of the existing graph objects and any subsequent output will be coerced to the desired class. Generally speaking, this corresponds to the output of \code{estimate.dag} and any corresponding \code{sparsebnPath} objects. Alternatively, users can \emph{manually} do this conversion on an object-by-object basis using the following methods:
\begin{itemize}
\item \code{to_igraph}: Conversion to and from \code{igraph} graphs from the \pkg{igraph} package;
\item \code{to_graphNEL}: Conversion to and from \code{graphNEL} graphs from the \pkg{graph} package;
\item \code{to_network}: Conversion to and from \code{network} graphs from the \pkg{network} package;
\item \code{to_bn}: Conversion to and from \code{bn} graphs from the \pkg{bnlearn} package.
\end{itemize}

\noindent
Each of these methods works on \code{sparsebnPath}, \code{sparsebnFit}, \code{edgeList} objects, in addition to any of the objects listed above.

Finally, the \pkg{sparsebn} package is compatible with the popular \pkg{Cytoscape} application \citep{shannon2003}. This is a standalone graphical interface for visualizing and analyzing complex networks. This is accomplished via the \code{openCytoscape} method, which leverages the \pkg{RCy3} package \citep{shannon2013} under the hood. In order to use this method, both \pkg{RCy3} and \pkg{Cytoscape} must be installed. For an example of this method in use, see Section~\ref{subsec:viz}.

\subsection{Installation}

\pkg{sparsebn} is an open-source package and is made freely available through CRAN. To install the latest stable version in \proglang{R},
\begin{Sinput}
R> install.packages("sparsebn")
\end{Sinput}

\noindent
For advanced users, the development versions can be downloaded directly from GitHub. Using \pkg{devtools}, the entire suite of packages can be installed via
\begin{Sinput}
R> devtools::install_github(c("itsrainingdata/sparsebnUtils/dev", 
+    "itsrainingdata/ccdrAlgorithm/dev", "gujyjean/discretecdAlgorithm/dev", 
+    "itsrainingdata/sparsebn/dev")) 
\end{Sinput}

\noindent
Note that before being released to CRAN, development versions may be unstable.

\section{Example: Cytometry data}
\label{sec:cyto}

To illustrate the use of this package, we will use the flow cytometry dataset from \citet{sachs2005} as a working example in this section. The original dataset consists of $n=7466$ observations of $p=11$ continuous variables corresponding to different proteins and phospholipids in human immune system cells, and each observation indicates the measured level of each biomolecule in a single cell under different experimental interventions. A network consisting of all well-established causal interactions between these molecules has been constructed based on biological experiments and literature. This network is frequently used as a benchmark to assess the accuracy of BN learning algorithms on real data.
Therefore, we refer to this as the consensus network in the sequel.

\begin{figure}[t]
\centering
\includegraphics[width=0.5\textwidth]{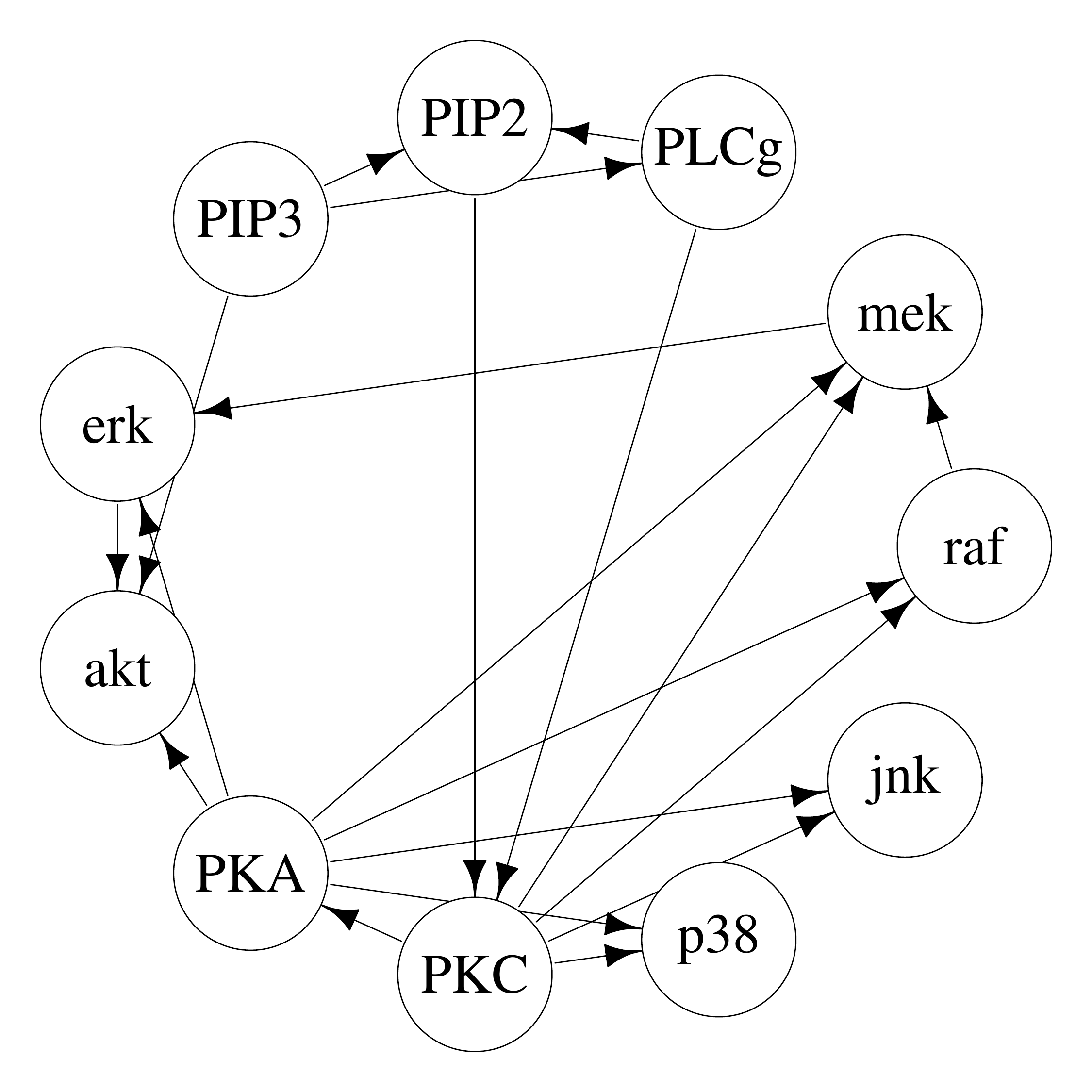}
\caption{The consensus cytometry network (11 nodes, 17 edges). A topological ordering of this network is PIP3$\,\prec\,$PLCg$\,\prec\,$PIP2$\,\prec\,$PKC$\,\prec\,$PKA$\,\prec\,$raf$\,\prec\,$mek$\,\prec\,$erk$\,\prec\,$akt$\,\prec\,$p38$\,\prec\,$jnk.}
\label{fig:cyto:network}
\end{figure}

The consensus network is visualized in Figure~\ref{fig:cyto:network}, in which a directed edge indicates that a change in the level of the parent will cause a change in the level of the child. This is a relatively small network which we use in order to keep the exposition simple. More examples, including a discrete version of this dataset and applications to large networks with hundreds or thousands of nodes, will be discussed in Section~\ref{sec:ex}.

First, we load this data:
\begin{Sinput}
R> library("sparsebn")
R> data("cytometryContinuous")
R> names("cytometryContinuous")
\end{Sinput}

\begin{Soutput}                          
  [1] "dag"  "data" "ivn" 
\end{Soutput}

\noindent
Note that this is not a \code{data.frame}, but instead a list of \proglang{R} objects that will be useful for this specific example. Each component of this list stores an important part of the experiment:
\begin{itemize}
\item \code{dag} is the consensus network with 11 nodes and 17 edges, as described above.
\item \code{data} is raw data collected from these experiments.
\item \code{ivn} is a list of interventions for each observation in the dataset.
\end{itemize}

\noindent
To illustrate the use of this package, the rest of this section describes the main steps to learning Bayesian networks from data: 1) Loading the data, 2) Learning the structure, 3) Incorporating prior knowledge, 4) Exploring the solution path and estimated networks, 5) Estimating the parameters, 6) Selecting the regularization parameter, and 7) Visualizing and assessing the output. 

\subsection{Loading data}
\label{subsec:load}

In order to distinguish different types of data---namely, experimental versus observational and continuous versus discrete---we use the \code{sparsebnData} class which wraps a \code{data.frame} into an object containing this auxiliary information. All of the methods implemented in \pkg{sparsebn} expect input as \code{sparsebnData}. 

To use this class, we need two important pieces of information: The raw data as a \code{data.frame}, and a list of interventions for each observation in the dataset. If the dataset does not contain any interventions, then the latter can be omitted. In order to create a \code{sparsebnData} object from the cytometry data, we first extract the necessary objects from \code{cytometryContinuous}:
\begin{Sinput}
R> cyto.raw <- cytometryContinuous$data
R> cyto.ivn <- cytometryContinuous$ivn
\end{Sinput}

\noindent
Now we can create the required \code{sparsebnData} object:
\begin{Sinput}
R> cyto.data <- sparsebnData(cyto.raw, type = "continuous", 
+    ivn = cyto.ivn)
\end{Sinput}

\noindent
Notice that we need to explicitly specify that the data is continuous (for discrete data, one would specify \code{type = "discrete"}). 

Finally, for discrete data, we may wish to manually specify the levels of each variable, which can be done using the \code{levels} argument. When this argument is omitted, we attempt to automatically infer the levels. Note that in doing so, however, levels which are missing from the data will not be recognized by the \code{sparsebnData} constructor.

\begin{Sinput}
R> print(cyto.data)
\end{Sinput}

\begin{Soutput}                          
       raf     mek  plc pip2  pip3  erk  akt  pka  pkc   p38   jnk
   1: 3.27 2.58022 2.18 2.91 4.074 1.89 2.83 6.03 2.83 3.804 3.689
   2: 3.58 2.80336 2.51 2.82 2.096 2.92 3.48 5.86 1.21 2.803 4.119
   3: 4.08 3.78646 2.68 2.32 2.565 2.70 3.48 6.00 2.43 3.463 2.970
   4: 4.29 4.41643 3.14 2.60 0.255 1.76 2.47 6.27 2.62 3.353 3.140
   5: 3.52 2.98568 1.65 2.28 3.211 3.05 3.83 5.72 1.54 3.246 4.398
  ---                                                             
7462: 3.89 2.51770 3.49 3.33 3.122 2.46 3.64 7.04 0.00 0.936 0.000
7463: 3.15 1.52823 2.88 3.10 2.701 3.89 4.21 6.83 0.00 2.284 0.000
7464: 3.34 1.50185 2.93 3.01 2.322 1.12 3.09 6.59 0.00 0.560 0.693
7465: 3.54 1.96009 1.75 3.03 2.715 3.47 3.72 6.70 3.80 7.231 0.892
7466: 3.42 0.00995 1.99 5.15 3.131 1.89 2.62 6.79 0.00 0.000 0.501

7466 total rows (7456 rows omitted)
Continuous data w/ interventions on 4863/7466 rows.
\end{Soutput}

\begin{Sinput}
R> summary(cyto.data)
\end{Sinput}

\begin{Soutput}                          
      raf            mek            plc            pip2     
 Min.   :0.00   Min.   :0.00   Min.   :0.00   Min.   :0.00  
 1st Qu.:3.43   1st Qu.:2.80   1st Qu.:2.24   1st Qu.:2.91  
 Median :3.99   Median :3.28   Median :2.80   Median :3.97  
 Mean   :4.09   Mean   :3.53   Mean   :2.88   Mean   :3.90  
 3rd Qu.:4.63   3rd Qu.:4.17   3rd Qu.:3.30   3rd Qu.:5.15  
 Max.   :8.44   Max.   :8.87   Max.   :8.73   Max.   :9.11  
      pip3           erk            akt            pka      
 Min.   :0.00   Min.   :0.00   Min.   :0.00   Min.   :0.00  
 1st Qu.:2.26   1st Qu.:2.14   1st Qu.:3.15   1st Qu.:5.62  
 Median :2.88   Median :2.84   Median :3.62   Median :6.11  
 Mean   :2.82   Mean   :2.75   Mean   :3.79   Mean   :5.83  
 3rd Qu.:3.49   3rd Qu.:3.47   3rd Qu.:4.28   3rd Qu.:6.62  
 Max.   :7.15   Max.   :7.85   Max.   :8.18   Max.   :9.09  
      pkc            p38            jnk      
 Min.   :0.00   Min.   :0.00   Min.   :0.00  
 1st Qu.:1.50   1st Qu.:2.96   1st Qu.:2.08  
 Median :2.54   Median :3.42   Median :2.91  
 Mean   :2.37   Mean   :3.53   Mean   :3.00  
 3rd Qu.:3.16   3rd Qu.:3.90   3rd Qu.:3.97  
 Max.   :7.38   Max.   :8.92   Max.   :8.46  

7466 total rows (7456 rows omitted)
Continuous data w/ interventions on 4863/7466 rows.
\end{Soutput}

\noindent
Note that some of the observations were not under intervention---such datasets with mixed observational and experimental samples are automatically handled by the methods in this package.

\subsection{Structure learning}
\label{subsec:learning}

To learn the structure of a Bayesian network from this data we use the \code{estimate.dag()} method, which runs the algorithm outlined in Section~\ref{subsec:algorithm}. To call this method using the default parameter settings, use:
\begin{Sinput}
R> cyto.learn <- estimate.dag(cyto.data)
R> print(cyto.learn)
\end{Sinput}

\begin{Soutput}
sparsebn Solution Path
 11 nodes
 7466 observations
 20 estimates for lambda in [0.8641, 86.406]
 Number of edges per solution: 0-1-6-8-13-15-15-19-22-21-26-33-35-36-38-38-
 41-43-46-50
\end{Soutput}

\begin{Sinput}
R> summary(cyto.learn)
\end{Sinput}

\begin{Soutput}
sparsebn Solution Path
 11 nodes
 7466 observations
 20 estimates for lambda in [0.8641, 86.406]
 Number of edges per solution: 0-1-6-8-13-15-15-19-22-21-26-33-35-36-38-38-
 41-43-46-50

   lambda nedge
1  86.406     0
2  67.808     1
3  53.213     6
4  41.759     8
5  32.771    13
6  25.717    15
7  20.182    15
8  15.838    19
9  12.429    22
10  9.754    21
11  7.654    26
12  6.007    33
13  4.714    35
14  3.699    36
15  2.903    38
16  2.278    38
17  1.788    41
18  1.403    43
19  1.101    46
20  0.864    50
\end{Soutput}

In addition to \code{data}, there are several optional parameters that can be passed to \code{estimate.dag}. The main arguments of interest are \code{lambdas} and \code{lambdas.length}, which allow the user to adjust the grid of regularization parameters $\lambda_{\max}>\lambda_{1}>\cdots>\lambda_{\min}$ used by the algorithms (Section~\ref{subsec:algorithm}). 

By default, \code{estimate.dag} produces a solution path of 20 estimates with the grid chosen adaptively to the data. This grid can be shortened or lengthened by specifying \code{lambdas.length}:
\begin{Sinput}
R> estimate.dag(cyto.data, lambdas.length = 50)
\end{Sinput}

\begin{Soutput}
sparsebn Solution Path
 11 nodes
 7466 observations
 50 estimates for lambda in [0.8641, 86.406]
 Number of edges per solution: 0-0-1-1-2-6-6-7-8-11-12-12-13-15-15-16-16-17-
 19-19-20-21-21-21-21-24-27-28-31-35-35-35-34-36-36-37-38-38-38-38-39-41-42-
 42-44-45-45-47-48-50
\end{Soutput}

For even more fine-tuning, the \code{lambdas} argument allows the user to explicitly input their own grid. For convenience we have included the \code{generate.lambdas} method, which provides a mechanism for generating grids of arbitrary lengths on either a linear or log scale. To generate a grid with a linear scale, use \code{scale = "linear"}:
\begin{Sinput}
R> cyto.lambdas <- generate.lambdas(lambda.max = 10, lambdas.ratio = 0.001, 
+    lambdas.length = 10, scale = "linear")
R> cyto.lambdas
\end{Sinput}

\begin{Soutput}
[1] 10.00  8.89  7.78  6.67  5.56  4.45  3.34  2.23  1.12  0.01
\end{Soutput}

To use a log scale, use \code{scale = "log"}:

\begin{Sinput}
R> cyto.lambdas <- generate.lambdas(lambda.max = 10, lambdas.ratio = 0.001, 
+    lambdas.length = 10, scale = "log")
R> cyto.lambdas
\end{Sinput}

\begin{Soutput}
[1] 10.0000  4.6416  2.1544  1.0000  0.4642  0.2154  0.1000  0.0464
[9]  0.0215  0.0100
\end{Soutput}

\noindent
This grid can also be generated manually, although this is not recommended. To run the algorithm using \code{cyto.lambdas} (output suppressed below):
\begin{Sinput}
R> estimate.dag(cyto.data, lambdas = cyto.lambdas)
\end{Sinput}

Another argument of interest is \code{edge.threshold}, which is another way to specify when the algorithm terminates. Specifically, if any point on the solution path contains an estimate with more than \code{edge.threshold} edges, the algorithm will terminate immediately and return what has been estimated up to that point. This makes our methods \emph{anytime algorithms}, in the sense that they can be interrupted at anytime while still producing valid output. This is convenient when running tests on very large graphs.

\subsection{Prior knowledge}

In some contexts, users may have prior knowledge regarding edges that must be present or absent from the network. For example, it may already be known that PIP3 regulates PIP2 (see Figure~\ref{fig:cyto:network}). In this case, estimation of the underlying network can be substantially improved by incorporating this information into the estimation procedure. With the \pkg{sparsebn} package, this can be done via \emph{whitelists} and \emph{blacklists}, which specify edges that must be present and absent, respectively. 

For example, to specify a known relationship between PIP3 and PIP2, we can create a whitelist as follows:
\begin{Sinput}
R> whitelist <- matrix(c("pip3", "pip2"), nrow = 1)
R> estimate.dag(cyto.data, whitelist = whitelist)
\end{Sinput}

\noindent
The \code{whitelist} argument should be a two-column matrix, where the first column stores parents and the second stores children (i.e., a from-to adjacency list):
\begin{Sinput}
R> whitelist
\end{Sinput}

\begin{Soutput}
     [,1]   [,2]  
[1,] "pip3" "pip2"
\end{Soutput}

Thus, this whitelist ensures that the edge PIP3$\to$PIP2 will be present in the final estimates.

Similarly, we can specify a blacklist, which stores edges that are known to be absent. For example, we can forbid any edges between RAF and MEK as follows:
\begin{Sinput}
R> blacklist <- rbind(c("raf", "jnk"), c("jnk", "raf"))
R> estimate.dag(cyto.data, blacklist = blacklist)
\end{Sinput}

\noindent
As with the whitelist, the blacklist should be a two-column matrix. Note that we specify \emph{both} directions RAF$\to$MEK and MEK$\to$RAF. If it is known that the direction can only go in one direction, then a single direction may be specified instead.

Blacklists are useful for specifying known root and leaf nodes in a Bayesian network. In the cytometry network, PIP3 is a root node (i.e., it has no parents). Thus, we can forbid any edges pointing \emph{into} PIP3. Similarly, JNK, P38, and AKT are leaf nodes (i.e., they have no children), so we can forbid any edges pointing \emph{away} from all three nodes. To specify this, we make use of the \code{specify.prior} function, which automatically builds an appropriate blacklist given the names of the root and leaf nodes. Any number of root and/or leaf nodes can be specified.

\begin{Sinput}
R> blacklist <- specify.prior(roots = "pip3", leaves = c("jnk", "p38", "akt"), 
+    nodes = names(cyto.data$data))
R> estimate.dag(cyto.data, blacklist = blacklist)
\end{Sinput}

Finally, whitelists and blacklists can be combined arbitrarily, as long as they are consistent in the sense that no edge appearing in the whitelist appears in the blacklist, and vice versa. This allows for a powerful specification of prior knowledge in learning networks from data.

\subsection{Solution paths}

The output of \code{estimate.dag} is a \code{sparsebnPath} object, which stores the entire solution path that is returned by the method.\footnote{This is similar in spirit to the \pkg{glasso} package \citep{friedman2014}, however, instead of storing each estimate as an \proglang{R} matrix we use the internal class \code{sparsebnFit}.} Since \code{sparsebnPath} objects also inherit from the \code{list} class in base \proglang{R}, we can inspect the first solution using ordinary \proglang{R} indexing. Note that for \code{sparsebnFit} objects, the \code{print} and \code{summary} methods are identical, so the output below is shown only once.

\begin{Sinput}
R> print(cyto.learn[[1]])
R> summary(cyto.learn[[1]])
\end{Sinput}

\begin{Soutput}
CCDr estimate
7466 observations
lambda = 86.4060183089118

DAG: 
<Empty graph on 11 nodes.>
\end{Soutput}

\noindent
The first estimate will always be the empty graph, which is a consequence of how we employ warm starts in the solution path. The third estimate, for example, shows a bit more structure:
\begin{Sinput}
R> print(cyto.learn[[3]])
R> summary(cyto.learn[[3]])
\end{Sinput}

\begin{Soutput}
CCDr estimate
7466 observations
lambda = 53.2129918008817

DAG: 
[raf]   mek   
[mek]    
[plc]    
[pip2]  plc   
[pip3]   
[erk]   akt   
[akt]    
[pka]   p38   
[pkc]    
[p38]   pkc   
[jnk]   pkc 
\end{Soutput}

\noindent
Each row in the output above corresponds to a child node---indicated by the square brackets---with its parents listed to the right without brackets. Formally, this is an adjacency list ordered by children.
For large graphs, explicit output of the parental structure as shown here is omitted by default, however, this behaviour can be overridden via the \code{maxsize} argument. Alternatively, we can retrieve the adjacency matrix, $\{I(\dagcompest_{ij}\ne 0)\}_{p\times p}$, for this estimate:
\begin{Sinput}
R> get.adjacency.matrix(cyto.learn[[3]])
\end{Sinput}

\begin{Soutput}
11 x 11 sparse Matrix of class "dgCMatrix"
                           
raf  . . . . . . . . . . .
mek  1 . . . . . . . . . .
plc  . . . 1 . . . . . . .
pip2 . . . . . . . . . . .
pip3 . . . . . . . . . . .
erk  . . . . . . . . . . .
akt  . . . . . 1 . . . . .
pka  . . . . . . . . . . .
pkc  . . . . . . . . . 1 1
p38  . . . . . . . 1 . . .
jnk  . . . . . . . . . . .
\end{Soutput}

\noindent
Note the use of the \pkg{Matrix} package \citep{bates2017}, which reduces the memory footprint on large graphs.

Finally, for large graphs, it may be desirable to inspect a subset of nodes, which can be done using the \code{show.parents} method:
\begin{Sinput}
R> show.parents(cyto.learn[[3]], c("raf", "pip2"))
\end{Sinput}

\begin{Soutput}
[raf]   mek   
[pip2]  plc  
\end{Soutput}

\subsection{Parameter estimation}
\label{subsec:inference}

It is important to note that the output of \code{estimate.dag} is a sequence of \emph{graphs}, i.e., no parameters (edge weights, variances, etc.) have been estimated at this stage. The next step is to estimate the values of the parameters associated with the underlying joint distribution. This is easy to do:
\begin{Sinput}
R> cyto.param <- estimate.parameters(cyto.learn, data = cyto.data)
\end{Sinput}

\noindent
The output is a list with each entry containing a component for the weighted adjacency matrix (\code{coefs}) and a diagonal matrix for the conditional variances (\code{vars}). For example, the coeffients of the third estimate in the solution path with six edges are (rounded to two decimal places):
\begin{Sinput}
R> cyto.param[[3]]$coefs
\end{Sinput}

\begin{Soutput}                          
11 x 11 sparse Matrix of class "dgCMatrix"
                                             
 [1,] .    . . .    . .     . .    . .   .   
 [2,] 1.05 . . .    . .     . .    . .   .   
 [3,] .    . . 1.26 . .     . .    . .   .   
 [4,] .    . . .    . .     . .    . .   .   
 [5,] .    . . .    . .     . .    . .   .   
 [6,] .    . . .    . .     . .    . .   .   
 [7,] .    . . .    . 0.725 . .    . .   .   
 [8,] .    . . .    . .     . .    . .   .   
 [9,] .    . . .    . .     . .    . 1.3 1.13
[10,] .    . . .    . .     . 1.36 . .   .   
[11,] .    . . .    . .     . .    . .   .   
\end{Soutput}

Similarly, we can inspect the estimated conditional variances:
\begin{Sinput}
R> Matrix::diag(cyto.param[[3]]$vars)
\end{Sinput}

\begin{Soutput}    
[1] 1.165 2.632 1.583 2.111 0.992 0.684 0.968 8.425 1.831 1.506 1.676
\end{Soutput}

Although the \code{vars} argument is a diagonal matrix, we have invoked the \code{diag} method above purely to save space.

For Gaussian data, we can also use Equation~\ref{eq:impliedcov} to estimate the implied covariance matrix for each solution in the solution path. This is implemented via the \code{get.covariance} method (see also \code{get.precision} for computing the precision, or inverse covariance, matrix). If the user is only interested in the covariance matrix (resp. precision matrix) and not the underlying DAG, then this extra step can be skipped by directly using the \code{estimate.covariance} method (resp. \code{estimate.precision}).

\subsection{Model selection}
\label{subsec:selection}

Unlike existing methods, the output of \code{estimate.dag} is a solution path with multiple estimates of increasing complexity, indexed by the regularization parameter. 
Thus, it is important to be able to pick out estimates for inspection and further exploration.
For \emph{ad hoc} exploration, the \code{select} method is useful: This allows one to select an estimate from a \code{sparsebnPath} object based on the number of edges, the regularization parameter $\lambda$, or the index $j$. When selecting by the number of edges or by $\lambda$, fuzzy matching is used by default so that the closest match is returned to within a given tolerance. Selecting by index is equivalent to subsetting as usual with the subset operator \code{`[[`}. To save space, the output of the following code is suppressed:

\begin{Sinput}
R> select(cyto.learn, edges = 8)  
R> select(cyto.learn, edges = 10) 
\end{Sinput}

In the first line above, an exact match is returned. In the second line, the closest match is returned since there is no graph with exactly 10 edges in the solution path.

\begin{Sinput}
R> select(cyto.learn, lambda = 41.75)
R> select(cyto.learn, lambda = 41.7) 
\end{Sinput}

In both of the above examples, the closest match is returned. 

\begin{Sinput}
R> select(cyto.learn, index = 4)
R> cyto.learn[[4]]              
\end{Sinput}

In the both lines above, an exact match is returned. Note that the second line is equivalent to the first.

For practical applications, one is often concerned with selecting an optimal value of $\lambda$. In high-dimensions, optimal selection of $\lambda$ for finite samples is an open problem, and past work has shown that both the prediction oracle and cross-validated choices perform poorly \citep{meinshausen2006,fu2013}. Instead, \citet{fu2013} suggest a practical method for selecting $\lambda$ based on a trade-off between the increase in log-likelihood and the increase in complexity between solutions. This method is implemented in \pkg{sparsebn} via the method \code{select.parameter}:

\begin{Sinput}
R> selected.lambda <- select.parameter(cyto.learn, cyto.data)
R> selected.lambda
\end{Sinput}

\begin{Soutput}
[1] 8
\end{Soutput}

\noindent
\code{select.parameter} returns the optimal index according to this method, in this case $j=8$, corresponding to a value of $\lambda\approx15.83806$ and an estimated network of 19 edges.

\subsection{Visualization}
\label{subsec:viz}

In order to visualize graphs estimated by the \pkg{sparsebn} package, we make use of the visualization capabilities of existing graph packages (see Section~\ref{subsec:compat}). By default, \pkg{sparsebn} uses the popular \pkg{igraph} package:
\begin{Sinput}
R> getPlotPackage()
\end{Sinput}

\begin{Soutput}                          
[1] "igraph"
\end{Soutput}

In addition to \pkg{igraph}, \pkg{sparsebn} is also compatible with the plotting features of the \pkg{graph} \citep[via \pkg{Rgraphviz},][]{hansen2016} and \pkg{network} packages. It is easy to change which package is used for plotting:
\begin{Sinput}
R> setPlotPackage("network")
R> getPlotPackage()
\end{Sinput}

\begin{Soutput}                          
[1] "network"
\end{Soutput}

\begin{Sinput}
R> setPlotPackage("graph")
R> getPlotPackage()
\end{Sinput}

\begin{Soutput}                          
[1] "graph"
\end{Soutput}

\begin{Sinput}
R> setPlotPackage("igraph")
R> getPlotPackage()
\end{Sinput}

\begin{Soutput}                          
[1] "igraph"
\end{Soutput}

\begin{figure}
\centering
\includegraphics[width=0.95\textwidth]{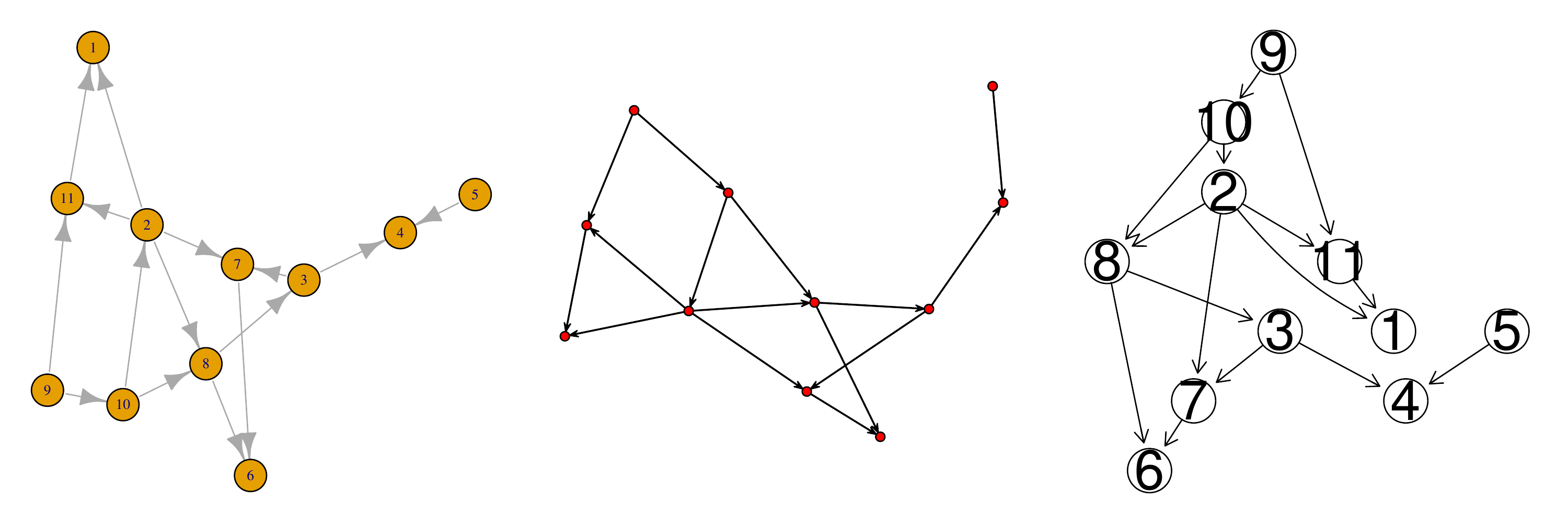}
\caption{Comparison of the default output from the three supported \proglang{R} packages for visualization: (left) \pkg{igraph}, (middle) \pkg{network}, (right) \pkg{graph}.}
\label{fig:cytocompare}
\end{figure}

\noindent
This allows the user complete flexibility over which \proglang{R} package is used for storing data and for visualizing data. In fact, \pkg{sparsebn} even allows one package to be used for visualization and a different package for storage. For comparison, the default output of each package is displayed in Figure~\ref{fig:cytocompare}. For large graphs, it is helpful to use a different set of defaults, provided in a separate method, \code{plotDAG}, which can be used for quick plotting (see, for example, Figure~\ref{fig:examplerun}).

Visualizing the full solution path is easy: Simply call \code{plot(cyto.learn)}. In order to ensure flexibility, we maintain all of the defaults used by the selected package for the \code{plot} method. It is easy, however, to customize the appearance of the plots if desired. For example, if we would like to compare the consensus cytometry network and the estimated network side by side, we can adjust the arguments to \code{plot} as follows:

\begin{Sinput}
R> plot(cytometryContinuous$dag,
+    layout = igraph::layout_(to_igraph(cytometryContinuous$dag), 
+                          igraph::in_circle()),
+    vertex.label = names(cytometryContinuous$dag),
+    vertex.size = 30,
+    vertex.label.color = gray(0),
+    vertex.color = gray(0.9),
+    edge.color = gray(0),
+    edge.arrow.size = 0.5)

R> plot(cyto.learn[[selected.lambda]],
+    layout = igraph::layout_(to_igraph(cytometryContinuous$dag), 
+                             igraph::in_circle()),
+    vertex.label = get.nodes(cyto.learn),
+    vertex.size = 30,
+    vertex.label.color = gray(0),
+    vertex.color = gray(0.9),
+    edge.color = gray(0),
+    edge.arrow.size = 0.5)
\end{Sinput}

\begin{figure}
\centering
\begin{subfigure}[b]{0.32\textwidth}
\includegraphics[width=\textwidth]{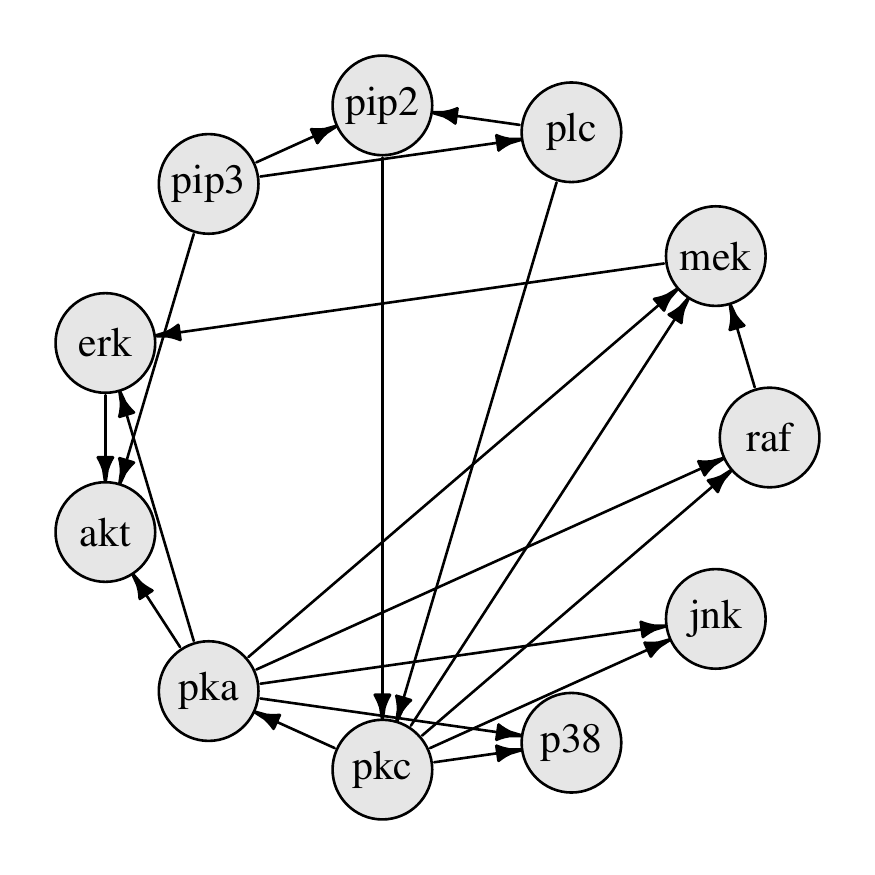}
\caption{Consensus network.}
\label{fig:cytoresults:consensus}
\end{subfigure}
\begin{subfigure}[b]{0.32\textwidth}
\includegraphics[width=\textwidth]{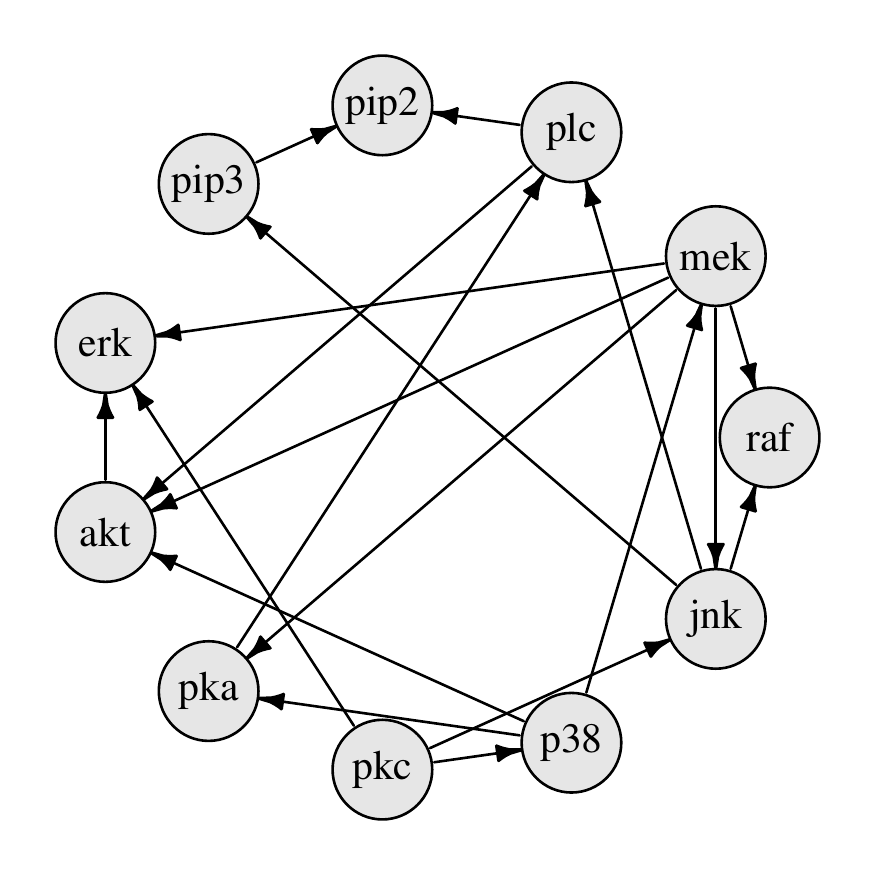}
\caption{Continuous network.}
\label{fig:cytoresults:cts}
\end{subfigure}
\begin{subfigure}[b]{0.32\textwidth}
\includegraphics[width=\textwidth]{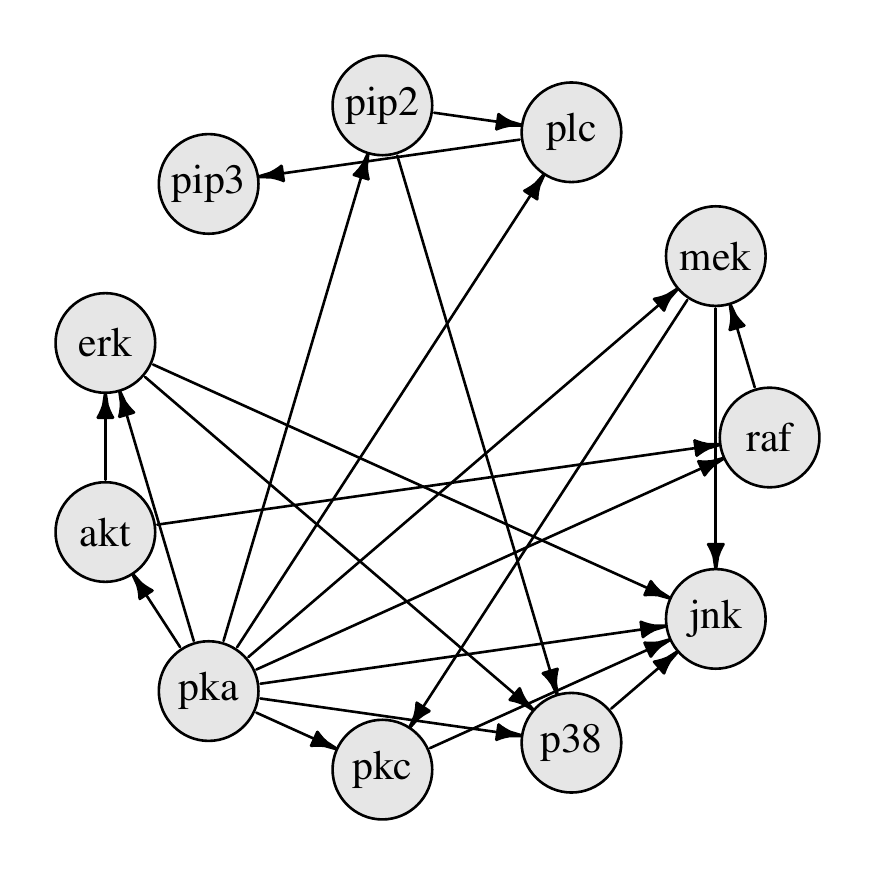}
\caption{Discrete network.}
\label{fig:cytoresults:disc}
\end{subfigure}
\caption{Cleaned up plots of (a) The consensus cytometry network, (b) The learned network based on continuous data, and (c) The learned network based on discretized data. Both learned networks were estimated using \code{estimate.dag}.}
\label{fig:cytoresults}
\end{figure}

\noindent
The output can be seen in Figures~\ref{fig:cytoresults:consensus} and~\ref{fig:cytoresults:cts}.

Finally, any object produced by \pkg{sparsebn} can be sent to the external application \pkg{Cytoscape}\footnote{\url{http://www.cytoscape.org/}} via the \code{openCytoscape} method. To use this method, \pkg{Cytoscape} must already be open and running in the background, and the \pkg{RCy3} package from \pkg{BioConductor} must also be installed.
\begin{Sinput}
R> openCytoscape(cytometryContinuous$dag)
R> openCytoscape(cyto.learn[[selected.lambda]])
\end{Sinput}

Calling either of these lines will creates a new network in \pkg{Cytoscape} application (outside of \proglang{R}), which can then be used for further analysis and investigation.

\section{Further examples}
\label{sec:ex}

In this section we provide three more examples of the functionality of \pkg{sparsebn}: 1) An example with discrete data, 2) A benchmark network from the Bayesian network repository, and 3) A gene expression dataset with $p=5000$ nodes.

\subsection{Discrete cytometry data}
\label{subsec:discrete}

In the previous section we used the cytometry network as an instructional example based on the original, continuous dataset. \cite{sachs2005} also provided a cleaned, discretized version of this dataset which can be used to illustrate how these methods apply to discrete data. After cleaning and pre-processing, the original continuous measurements were binned into one of three levels: \emph{low} = 0, \emph{medium} = 1, or \emph{high} = 2. Due to the pre-processing, the discrete data contains fewer observations ($n=5400$) compared to the raw, continuous data.

To use this data, we start by loading it as usual:

\begin{Sinput}
R> library("sparsebn")
R> data("cytometryDiscrete")
\end{Sinput}

\noindent
The code to estimate this graph is essentially the same as in Section~\ref{sec:cyto}. For completeness, we present only the essential steps here. As before, the first step is to pass the data into a \code{sparsebnData} object:
\begin{Sinput}
R> cyto.data <- sparsebnData(cytometryDiscrete$data, type = "discrete",
+    ivn = cytometryDiscrete$ivn)
R> cyto.data
\end{Sinput}

\begin{Soutput}
      raf mek plc pip2 pip3 erk akt pka pkc p38 jnk
   1:   0   0   0    1    2   1   0   2   0   1   0
   2:   0   0   0    0    2   2   1   2   0   1   0
   3:   0   0   1    1    2   1   0   2   1   0   0
   4:   0   0   0    0    2   1   0   2   0   2   0
   5:   0   0   0    0    2   1   0   2   0   0   0
  ---                                              
5396:   0   0   0    0    1   1   0   1   1   0   0
5397:   0   0   0    0    0   1   1   0   0   1   1
5398:   0   0   0    0    1   1   0   1   1   0   0
5399:   0   1   0    0    0   0   0   1   1   0   0
5400:   1   1   0    0    1   1   0   1   1   1   0

5400 total rows (5390 rows omitted)
Discrete data w/ interventions on 3600/5400 rows.
\end{Soutput}

\noindent
One of the main purposes of the \code{sparsebnData} class is to encode all of the necessary information needed to run the main algorithms; now that the data has been converted into a \code{sparsebnData} object, the user will notice almost no difference between the code in Section~\ref{sec:cyto} and the sequel.

To estimate a DAG based on the discrete data, use \code{estimate.dag}:
\begin{Sinput}
R> cyto.learn <- estimate.dag(cyto.data)
R> cyto.learn
\end{Sinput}

\begin{Soutput}
sparsebn Solution Path
 11 nodes
 5400 observations
 8 estimates for lambda in [1.791, 9.7712]
 Number of edges per solution: 0-6-9-13-15-21-30-38
\end{Soutput}

It is easy to adjust \code{lambdas} and \code{lambdas.length} as before. Note also the difference between the number of solutions here and in Section~\ref{subsec:learning}; this is a consequence of the stopping criterion used for discrete data (see also Section~\ref{subsec:algorithm}). For comparison with the network selected in Section~\ref{subsec:selection} which had 19 edges, we use \code{select} to choose the estimate that is closest in complexity; the closest such network in the present case has 21 edges and is shown in Figure~\ref{fig:cytoresults:disc}. The code to reproduce this figure is below:
\begin{Sinput}
R> plot(select(cyto.learn, edges = 19),
+    layout = igraph::layout_(to_igraph(cytometryContinuous$dag), 
+                             igraph::in_circle()),
+    vertex.label = get.nodes(cyto.learn),
+    vertex.size = 30,
+    vertex.label.color = gray(0),
+    vertex.color = gray(0.9),
+    edge.color = gray(0),
+    edge.arrow.size = 0.5)
\end{Sinput}

To estimate the parameters associated with the multi-logit model, use \code{estimate.parameters}:
\begin{Sinput}
R> cyto.param <- estimate.parameters(cyto.learn, data = cyto.data)
\end{Sinput}

As the parameter space for the multi-logit model is much larger than for the Gaussian model, the output is much more complicated. For example, the node RAF has a single parent PKA in the DAG selected above, and so the parameter space for this node is a $2\times 3$ matrix:

\begin{Sinput}
R> cyto.param[[5]][["raf"]]
\end{Sinput}

\begin{Soutput}
      (Intercept)      pka_1     pka_2
    1   0.4219942 -0.9457959 -2.282747
    2   1.8258925 -3.6595227 -5.034717
\end{Soutput}

\noindent
For each extra parent, there will be two more columns added to this matrix, owing to the fact that each variable has three levels. There is one such matrix for each node (see Section~\ref{subsec:paraminference} for details on the estimated parameters of the multi-logit model). 

\subsection{The pathfinder network}
\label{subsec:pathfinder}

In order to illustrate this package on a larger network, we will reconstruct the pathfinder network \citep{heckerman1992}. The pathfinder network has 109 nodes and 195 edges and is part of the Bayesian network repository,\footnote{\url{http://www.bnlearn.com/bnrepository/}} a centralized repository of benchmark networks for testing structure learning algorithms.

We first load the dataset:
\begin{Sinput}
R> data("pathfinder")
R> dat <- sparsebnData(pathfinder$data, type = "c")
\end{Sinput}

\noindent
The data was generated from a Gaussian SEM with $\beta_{ij}=1$ whenever $\beta_{ij}\ne 0$ and $\omega_{j}^{2}=1$ for each $j$. By Equation~\ref{eq:impliedcov}, we were able to compute the implied covariance matrix and use \pkg{mvtnorm} to generate samples from this distribution \citep{genz2017}. For this example, $n=1000$ samples were drawn.

For illustrative purposes, we will estimate a longer solution path with 50 DAGs:
\begin{Sinput}
R> nn <- num.samples(dat)
R> lambdas <- generate.lambdas(sqrt(nn), 0.05, lambdas.length = 50, 
+    scale = "linear")
R> dags <- estimate.dag(data = dat, lambdas = lambdas, edge.threshold = 500, 
+    verbose = FALSE)
R> dags
\end{Sinput}

\begin{Soutput}                          
 109 nodes
 1000 observations
 50 estimates for lambda in [1.5811, 31.6228]
 Number of edges per solution: 0-17-22-29-34-37-39-50-58-59-63-63-64-64-
 65-78-103-108-108-108-108-108-108-108-108-108-108-108-108-108-113-115-
 119-121-121-121-124-130-135-137-139-144-153-166-180-189-206-219-249-370
\end{Soutput}

\noindent
The choice of $\lambda_{\max}=n^{1/2}$ in \code{generate.lambdas} is important as it guarantees that the first solution will be the empty graph \citep[see][Section~5.3]{aragam2015}. We use a linear scale for the grid and we set \code{edge.threshold = 500} in order to terminate the algorithm early if any estimate has more than 500 edges (although note that in this case this constraint never becomes active).

Furthermore, there are no difficulties if the data is high-dimensional ($p>n$). Let us extract the first 50 rows of $\samplemat$, so that $p=109>n=50$:
\begin{Sinput}
R> dat <- sparsebnData(pathfinder$data[1:50, ], type = "c")
R> nn <- num.samples(dat)
R> lambdas <- generate.lambdas(sqrt(nn), 0.05, lambdas.length = 50, 
+    scale = "linear")
R> dags <- estimate.dag(data = dat, lambdas = lambdas, edge.threshold = 500, 
+    verbose = FALSE)
R> dags
\end{Sinput}

\begin{Soutput}                          
sparsebn Solution Path
 109 nodes
 50 observations
 43 estimates for lambda in [1.3132, 7.0711]
 Number of edges per solution: 0-16-21-23-28-39-39-42-44-49-53-58-63-64-67-
 69-71-80-81-84-88-92-99-102-107-111-116-116-118-123-127-134-142-149-161-
 173-193-217-229-262-311-382-477
\end{Soutput}

\noindent
Note that in this case the algorithm terminates just shy of the full 50 estimates, since the 44th solution apparently exceeds the threshold of 500 edges. Finally, for this high-dimensional example, there is no noticeable slowdown compared to the previous case with $n=1000$.

\subsection{Large networks}

To conclude the examples, we finish with an application to the estimation of very large networks using gene expression data collected from $129$ late-onset Alzheimer's disease (LOAD) patients and 101 healthy patients \citep{zhang2013}. The data consists of post-mortem whole-genome gene-expression profiles from autopsied brain tissues and in particular constitutes a purely observational sample. In total there are three datasets, each corresponding to tissue samples collected from a different brain region: 1) Dorsolateral prefrontal cortex (PFC), 2) Visual cortex (VC) and 3) Cerebellum (CR). Since ground truth networks have not been established for these datasets, the main purpose of this example is to illustrate how \pkg{sparsebn} scales to real world, high-dimensional data with thousands of variables. 

Each dataset contains expression measurements from $39,280$ probes, of which we extracted the first $p=5,000$ columns for use in this example. For this example, we focused on LOAD patients only, so $n=129$. We then used \code{estimate.dag} to learn a solution path from each dataset, and kept track of the time. In our experiments, it took a little over four minutes for our package to estimate each solution path, assuming a maximum of 5000 edges. For example, assuming the output has been stored in a variable called \code{dags}, the resulting output for the PFC data is\footnote{Code to reproduce this experiment can be found at \url{https://github.com/itsrainingdata/sparsebn-reproduce}.}
\begin{Sinput}
R> dags
\end{Sinput}

\begin{Soutput}                          
sparsebn Solution Path
 5000 nodes
 129 observations
 15 estimates for lambda in [2.9973, 11.3578]
 Number of edges per solution: 0-86-465-1326-2344-3190-3850-4465-4903-4967-
 4978-4988-4989-4992-4997
\end{Soutput}

The output for the remaining two datasets is similar, consisting of 15 DAGs for the VC data and 16 DAGs for the CR data.

Some comments on the timing are in order. The \code{edge.threshold} argument was used here to terminate the algorithm at 5000 edges---in practice, of course, we do not know the true number of edges and so this is meant to be purely illustrative. Moreover, since this parameter can be set arbitrarily high, the algorithm can in principle run for arbitrarily long. In fact, one of the advantages of our methods is that they can be interrupted at anytime, bearing in mind that the result may be a \emph{sub}graph of the true graph. The \code{edge.threshold} parameter allows the user to incorporate prior knowledge of the sparsity (or lack thereof) of the underlying graph.

\section{Conclusion}
\label{sec:conc}

\pkg{sparsebn} is a fast, modern package for learning the structure of sparse Bayesian networks. By leveraging recent trends in nonconvex optimization, sparse regularization, and causal inference, we are able to scale structure learning to problems containing high-dimensional data with thousands of variables and experimental interventions. This fills a gap within existing software packages for learning Bayesian networks, which already provide excellent coverage for traditional problems with large samples and without interventions. All of the code to reproduce the results presented here is available on GitHub, along with the source code of the \pkg{sparsebn} package, which is available on CRAN.

Given the nonconvex nature of the minimization problems here, one future direction is to incorporate stochastic optimization into our package to enhance its global search ability. For example, stochastic gradient descent may be used in the algorithm for discrete Bayesian networks, which will reduce the computational complexity as well. We are also interested in developing divide-and-conquer strategies for ultra-large graphs, say on the scale of $10^5$ nodes. Along these lines, we are also exploring parallel and distributed implementations of these algorithms. Finally, our package can be combined with various post-learning functions for multi-stage learning of causal networks. Given the DAG learned from \pkg{sparsebn}, one may further infer causal relations in a subgraph of interest by making additional causal assumptions or with new experimental data.

\section*{Acknowledgements} 

This work was supported by NSF grants IIS-1546098 and DMS-1055286 to Q.Z. The authors thank Dacheng Zhang for helpful discussions and computational assistance.

\bibliography{sparsebn-jss}

\begin{thebibliography}{82}
\newcommand{\enquote}[1]{``#1''}
\providecommand{\natexlab}[1]{#1}
\providecommand{\url}[1]{\texttt{#1}}
\providecommand{\urlprefix}{URL }
\expandafter\ifx\csname urlstyle\endcsname\relax
  \providecommand{\doi}[1]{doi:\discretionary{}{}{}#1}\else
  \providecommand{\doi}{doi:\discretionary{}{}{}\begingroup
  \urlstyle{rm}\Url}\fi
\providecommand{\eprint}[2][]{\url{#2}}

\bibitem[{Aragam \emph{et~al.}(2016)Aragam, Amini, and Zhou}]{aragam2016}
Aragam B, Amini AA, Zhou Q (2016).
\newblock \enquote{Learning Directed Acyclic Graphs with Penalized
  Neighbourhood Regression.}
\newblock \emph{Submitted}, \textbf{arXiv:1511.08963}.

\bibitem[{Aragam and Zhou(2015)}]{aragam2015}
Aragam B, Zhou Q (2015).
\newblock \enquote{Concave Penalized Estimation of Sparse {G}aussian {B}ayesian
  Networks.}
\newblock \emph{Journal of Machine Learning Research}, \textbf{16}, 2273--2328.

\bibitem[{Bates and Maechler(2017)}]{bates2017}
Bates D, Maechler M (2017).
\newblock \emph{Matrix: Sparse and Dense Matrix Classes and Methods}.
\newblock R package version 1.2-8,
  \urlprefix\url{https://CRAN.R-project.org/package=Matrix}.

\bibitem[{Boettcher and Dethlefsen(2003)}]{boettcher2003}
Boettcher S, Dethlefsen C (2003).
\newblock \enquote{\pkg{deal}: A Package for Learning Bayesian Networks.}
\newblock \emph{Journal of Statistical Software}, \textbf{8}(1), 1--40.
\newblock ISSN 1548-7660.
\newblock \doi{10.18637/jss.v008.i20}.
\newblock
  \urlprefix\url{https://www.jstatsoft.org/index.php/jss/article/view/v008i20}.

\bibitem[{Bouckaert(1993)}]{Bouckaert93prob}
Bouckaert RR (1993).
\newblock \enquote{Probabilistic Network Construction Using the Minimum
  Description Length Principle.}
\newblock In \emph{Symbolic and Quantitative Approaches to Reasoning and
  Uncertainty: European Conference ECSQARU '93}, volume 747 of \emph{Lecture
  Notes in Computer Science}, pp. 41--48. Springer-Verlag.

\bibitem[{B{\"u}hlmann \emph{et~al.}(2014)B{\"u}hlmann, Kalisch, and
  Meier}]{buhlmann2014bio}
B{\"u}hlmann P, Kalisch M, Meier L (2014).
\newblock \enquote{High-Dimensional Statistics with a View Toward Applications
  in Biology.}
\newblock \emph{Annual Review of Statistics and Its Application}, \textbf{1}.

\bibitem[{Buntine(1991)}]{Buntine91}
Buntine W (1991).
\newblock \enquote{Theory Refinement on {B}ayesian Networks.}
\newblock In \emph{Proceedings of the Seventh Annual Conference on Uncertainty
  in Artificial Intelligence}, pp. 52--60. Morgan Kaufmann.

\bibitem[{Butts(2008)}]{butts2008}
Butts C (2008).
\newblock \enquote{\pkg{network}: A Package for Managing Relational Data in
  \proglang{R}.}
\newblock \emph{Journal of Statistical Software}, \textbf{24}(1), 1--36.
\newblock ISSN 1548-7660.
\newblock \doi{10.18637/jss.v024.i02}.
\newblock
  \urlprefix\url{https://www.jstatsoft.org/index.php/jss/article/view/v024i02}.

\bibitem[{Chen \emph{et~al.}(2015)Chen, Wang, Chen, Liu, Tang, Mao, Li, Lin,
  Sun, and Ma}]{chen2015}
Chen YP, Wang ZX, Chen L, Liu X, Tang LL, Mao YP, Li WF, Lin AH, Sun Y, Ma J
  (2015).
\newblock \enquote{A Bayesian Network Meta-Analysis Comparing Concurrent
  Chemoradiotherapy Followed by Adjuvant Chemotherapy, Concurrent
  Chemoradiotherapy Alone and Radiotherapy Alone in Patients with
  Locoregionally Advanced Nasopharyngeal Carcinoma.}
\newblock \emph{Annals of Oncology}, \textbf{26}(1), 205--211.

\bibitem[{Chickering and Heckerman(1997)}]{Chickering97}
Chickering DM, Heckerman D (1997).
\newblock \enquote{Efficient Approximations for the Marginal Likelihood of
  {B}ayesian Networks with Hidden Variables.}
\newblock \emph{Machine Learning}, \textbf{29}, 181--212.

\bibitem[{Chickering \emph{et~al.}(2004)Chickering, Heckerman, and
  Meek}]{chickering2004}
Chickering DM, Heckerman D, Meek C (2004).
\newblock \enquote{Large-Sample Learning of {B}ayesian Networks is {NP}-hard.}
\newblock \emph{The Journal of Machine Learning Research}, \textbf{5},
  1287--1330.

\bibitem[{Colombo \emph{et~al.}(2012)Colombo, Maathuis, Kalisch, and
  Richardson}]{colombo2012}
Colombo D, Maathuis MH, Kalisch M, Richardson TS (2012).
\newblock \enquote{Learning High-Dimensional Directed Acyclic Graphs with
  Latent and Selection Variables.}
\newblock \emph{The Annals of Statistics}, \textbf{40}(1), 294--321.

\bibitem[{Cooper and Herskovits(1992)}]{Cooper92}
Cooper GF, Herskovits E (1992).
\newblock \enquote{A {B}ayesian Method for the Induction of Probabilistic
  Networks from Data.}
\newblock \emph{Machine Learning}, \textbf{9}, 309--347.

\bibitem[{Cooper and Yoo(1999)}]{cooper1999causal}
Cooper GF, Yoo C (1999).
\newblock \enquote{Causal Discovery from a Mixture of Experimental and
  Observational Data.}
\newblock In \emph{Proceedings of the Fifteenth conference on Uncertainty in
  artificial intelligence}, pp. 116--125. Morgan Kaufmann Publishers Inc.

\bibitem[{Csardi and Nepusz(2006)}]{csardi2006}
Csardi G, Nepusz T (2006).
\newblock \enquote{The \pkg{igraph} Software Package for Complex Network
  Research.}
\newblock \emph{InterJournal}, \textbf{Complex Systems}, 1695.
\newblock \urlprefix\url{http://igraph.org}.

\bibitem[{Cussens \emph{et~al.}(2017)Cussens, Haws, and
  Studen{\`y}}]{cussens2017}
Cussens J, Haws D, Studen{\`y} M (2017).
\newblock \enquote{Polyhedral Aspects of Score Equivalence in Bayesian Network
  Structure Learning.}
\newblock \emph{Mathematical Programming}, \textbf{164}(1-2), 285--324.

\bibitem[{Dejaeger \emph{et~al.}(2013)Dejaeger, Verbraken, and
  Baesens}]{dejaeger2013}
Dejaeger K, Verbraken T, Baesens B (2013).
\newblock \enquote{Toward Comprehensible Software Fault Prediction Models Using
  Bayesian Network Classifiers.}
\newblock \emph{IEEE Transactions on Software Engineering}, \textbf{39}(2),
  237--257.

\bibitem[{Dobson and Barnett(2008)}]{dobson2008}
Dobson AJ, Barnett A (2008).
\newblock \emph{An Introduction to Generalized Linear Models}.
\newblock CRC press.

\bibitem[{Eddelbuettel(2013)}]{eddelbuettel2013}
Eddelbuettel D (2013).
\newblock \emph{Seamless \proglang{R} and \proglang{C++} Integration with
  \pkg{Rcpp}}.
\newblock Springer-Verlag, New York.
\newblock ISBN 978-1-4614-6867-7.

\bibitem[{Eddelbuettel and Fran\c{c}ois(2011)}]{eddelbuettel2011}
Eddelbuettel D, Fran\c{c}ois R (2011).
\newblock \enquote{\pkg{Rcpp}: Seamless \proglang{R} and \proglang{C++}
  Integration.}
\newblock \emph{Journal of Statistical Software}, \textbf{40}(8), 1--18.
\newblock \urlprefix\url{http://www.jstatsoft.org/v40/i08/}.

\bibitem[{Ellis and Wong(2008)}]{Ellis08}
Ellis B, Wong WH (2008).
\newblock \enquote{Learning Causal {B}ayesian Network Structures from
  Experimental Data.}
\newblock \emph{Journal of the American Statistical Association}, \textbf{103},
  778--789.

\bibitem[{Fan and Li(2001)}]{fan2001}
Fan J, Li R (2001).
\newblock \enquote{Variable Selection via Nonconcave Penalized Likelihood and
  its Oracle Properties.}
\newblock \emph{Journal of the American Statistical Association},
  \textbf{96}(456), 1348--1360.

\bibitem[{Friedman \emph{et~al.}(2007)Friedman, Hastie, H{\"o}fling, and
  Tibshirani}]{friedman2007}
Friedman J, Hastie T, H{\"o}fling H, Tibshirani R (2007).
\newblock \enquote{Pathwise Coordinate Optimization.}
\newblock \emph{The Annals of Applied Statistics}, \textbf{1}(2), 302--332.

\bibitem[{Friedman \emph{et~al.}(2008)Friedman, Hastie, and
  Tibshirani}]{friedman2008}
Friedman J, Hastie T, Tibshirani R (2008).
\newblock \enquote{Sparse Inverse Covariance Estimation with the {G}raphical
  {L}asso.}
\newblock \emph{Biostatistics}, \textbf{9}(3), 432--441.

\bibitem[{Friedman \emph{et~al.}(2010)Friedman, Hastie, and
  Tibshirani}]{friedman2010}
Friedman J, Hastie T, Tibshirani R (2010).
\newblock \enquote{Regularization Paths for Generalized Linear Models via
  Coordinate Descent.}
\newblock \emph{Journal of statistical software}, \textbf{33}(1), 1.

\bibitem[{Friedman \emph{et~al.}(2014)Friedman, Hastie, and
  Tibshirani}]{friedman2014}
Friedman J, Hastie T, Tibshirani R (2014).
\newblock \emph{\pkg{glasso}: Graphical Lasso - Estimation of Gaussian
  Graphical Models}.
\newblock R package version 1.8,
  \urlprefix\url{https://CRAN.R-project.org/package=glasso}.

\bibitem[{Fu and Zhou(2013)}]{fu2013}
Fu F, Zhou Q (2013).
\newblock \enquote{Learning Sparse Causal {G}aussian Networks With Experimental
  Intervention: {R}egularization and Coordinate Descent.}
\newblock \emph{Journal of the American Statistical Association},
  \textbf{108}(501), 288--300.

\bibitem[{G{\'a}mez \emph{et~al.}(2011)G{\'a}mez, Mateo, and
  Puerta}]{gamez2011learning}
G{\'a}mez JA, Mateo JL, Puerta JM (2011).
\newblock \enquote{Learning Bayesian Networks by Hill Climbing: Efficient
  Methods Based on Progressive Restriction of the Neighborhood.}
\newblock \emph{Data Mining and Knowledge Discovery}, \textbf{22}(1-2),
  106--148.

\bibitem[{Gao and Cui(2015)}]{gao2015}
Gao B, Cui Y (2015).
\newblock \enquote{Learning Directed Acyclic Graphical Structures with
  Genetical Genomics Data.}
\newblock \emph{Bioinformatics}, p. btv513.

\bibitem[{Garvey \emph{et~al.}(2015)Garvey, Carnovale, and
  Yeniyurt}]{garvey2015}
Garvey MD, Carnovale S, Yeniyurt S (2015).
\newblock \enquote{An Analytical Framework for Supply Network Risk Propagation:
  A Bayesian Network Approach.}
\newblock \emph{European Journal of Operational Research}, \textbf{243}(2),
  618--627.

\bibitem[{Gentleman \emph{et~al.}(2016)Gentleman, Whalen, Huber, and
  Falcon}]{gentleman2016}
Gentleman R, Whalen E, Huber W, Falcon S (2016).
\newblock \emph{\pkg{graph}: A Package to Handle Graph Data Structures}.
\newblock R package version 1.50.0,
  \urlprefix\url{https://CRAN.R-project.org/package=graph}.

\bibitem[{Genz \emph{et~al.}(2017)Genz, Bretz, Miwa, Mi, Leisch, Scheipl, and
  Hothorn}]{genz2017}
Genz A, Bretz F, Miwa T, Mi X, Leisch F, Scheipl F, Hothorn T (2017).
\newblock \emph{\pkg{mvtnorm}: Multivariate Normal and t Distributions}.
\newblock R package version 1.0-6,
  \urlprefix\url{https://CRAN.R-project.org/package=mvtnorm}.

\bibitem[{Gu \emph{et~al.}(2018)Gu, Fu, and Zhou}]{gu2018}
Gu J, Fu F, Zhou Q (2018).
\newblock \enquote{Penalized Estimation of Directed Acyclic Graphs From
  Discrete Data.}
\newblock \emph{Statistics and Computing}, \textbf{DOI:
  10.1007/s11222-018-9801-y}.

\bibitem[{Hansen \emph{et~al.}(2016)Hansen, Gentry, Long, Gentleman, Falcon,
  Hahne, and Sarkar}]{hansen2016}
Hansen KD, Gentry J, Long L, Gentleman R, Falcon S, Hahne F, Sarkar D (2016).
\newblock \emph{\pkg{Rgraphviz}: Provides Plotting Capabilities for R Graph
  Objects}.
\newblock R package version 2.16.0,
  \urlprefix\url{https://CRAN.R-project.org/package=Rgraphviz}.

\bibitem[{Heckerman \emph{et~al.}(1995)Heckerman, Geiger, and
  Chickering}]{Heckerman95}
Heckerman D, Geiger D, Chickering DM (1995).
\newblock \enquote{Learning {B}ayesian Networks: The Combination of Knowledge
  and Statistical Data.}
\newblock \emph{Machine Learning}, \textbf{20}, 197--243.

\bibitem[{Heckerman \emph{et~al.}(1992)Heckerman, Horvitz, and
  Nathwani}]{heckerman1992}
Heckerman D, Horvitz E, Nathwani B (1992).
\newblock \enquote{Toward Normative Expert Systems: Part I, the Pathfinder
  Project. Knowledge Systems Laboratory, Medical Computer Science.}

\bibitem[{Herskovits and Cooper(1990)}]{Herskovits90}
Herskovits E, Cooper G (1990).
\newblock \enquote{Kutat\'{o}: An Entropy-Driven System for Construction of
  Probabilistic Expert Systems from Databases.}
\newblock In \emph{Proceedings of the Sixth Annual Conference on Uncertainty in
  Artificial Intelligence}, pp. 54--62.

\bibitem[{H{\o}jsgaard(2012)}]{hojsgaard2012}
H{\o}jsgaard S (2012).
\newblock \enquote{Graphical Independence Networks with the \pkg{gRain} Package
  for \proglang{R}.}
\newblock \emph{Journal of Statistical Software}, \textbf{46}(1), 1--26.
\newblock ISSN 1548-7660.
\newblock \doi{10.18637/jss.v046.i10}.
\newblock
  \urlprefix\url{https://www.jstatsoft.org/index.php/jss/article/view/v046i10}.

\bibitem[{Isci \emph{et~al.}(2014)Isci, Dogan, Ozturk, and Otu}]{isci2015}
Isci S, Dogan H, Ozturk C, Otu HH (2014).
\newblock \enquote{Bayesian Network Prior: Network Analysis of Biological Data
  Using External Knowledge.}
\newblock \emph{Bioinformatics}, \textbf{30}(6), 860--867.

\bibitem[{Jones \emph{et~al.}(2012)Jones, Buchan, Cozzetto, and
  Pontil}]{jones2012}
Jones DT, Buchan DW, Cozzetto D, Pontil M (2012).
\newblock \enquote{PSICOV: Precise Structural Contact Prediction Using Sparse
  Inverse Covariance Estimation on Large Multiple Sequence Alignments.}
\newblock \emph{Bioinformatics}, \textbf{28}(2), 184--190.

\bibitem[{Kalisch \emph{et~al.}(2012)Kalisch, M{\"a}chler, Colombo, Maathuis,
  and B{\"u}hlmann}]{kalisch2012}
Kalisch M, M{\"a}chler M, Colombo D, Maathuis MH, B{\"u}hlmann P (2012).
\newblock \enquote{Causal Inference Using Graphical Models with the
  \proglang{R} Package \pkg{pcalg}.}
\newblock \emph{Journal of Statistical Software}, \textbf{47}(11), 1--26.
\newblock \urlprefix\url{https://www.jstatsoft.org/article/view/v047i11}.

\bibitem[{Koller and Friedman(2009)}]{koller2009}
Koller D, Friedman N (2009).
\newblock \emph{Probabilistic Graphical Models: Principles and Techniques}.
\newblock MIT press.

\bibitem[{Lam and Bacchus(1994)}]{Lam94}
Lam W, Bacchus F (1994).
\newblock \enquote{Learning {B}ayesian Belief Networks: An Approach Based on
  the {MDL} Principle.}
\newblock \emph{Computational Intelligence}, \textbf{10}, 269--293.

\bibitem[{Lauritzen(1996)}]{lauritzen1996}
Lauritzen SL (1996).
\newblock \emph{Graphical Models}.
\newblock Oxford University Press.

\bibitem[{Masegosa \emph{et~al.}(2017)Masegosa, Mart{\'\i}nez, Ramos-L{\'o}pez,
  Caba{\~n}as, Salmer{\'o}n, Nielsen, Langseth, and Madsen}]{masegosa2017}
Masegosa AR, Mart{\'\i}nez AM, Ramos-L{\'o}pez D, Caba{\~n}as R, Salmer{\'o}n
  A, Nielsen TD, Langseth H, Madsen AL (2017).
\newblock \enquote{AMIDST: a Java Toolbox for Scalable Probabilistic Machine
  Learning.}
\newblock \emph{arXiv preprint arXiv:1704.01427}.

\bibitem[{Mazumder \emph{et~al.}(2011)Mazumder, Friedman, and
  Hastie}]{mazumder2011}
Mazumder R, Friedman JH, Hastie T (2011).
\newblock \enquote{Sparse{N}et: {C}oordinate Descent with Nonconvex Penalties.}
\newblock \emph{Journal of the American Statistical Association},
  \textbf{106}(495), 1125--1138.

\bibitem[{Meganck \emph{et~al.}(2006)Meganck, Leray, and
  Manderick}]{meganck2006learning}
Meganck S, Leray P, Manderick B (2006).
\newblock \enquote{Learning Causal Bayesian Networks from Observations and
  Experiments: A Decision Theoretic Approach.}
\newblock In \emph{Modeling Decisions for Artificial Intelligence}, pp. 58--69.
  Springer-Verlag.

\bibitem[{Meinshausen and B{\"u}hlmann(2006)}]{meinshausen2006}
Meinshausen N, B{\"u}hlmann P (2006).
\newblock \enquote{High-Dimensional Graphs and Variable Selection with the
  {L}asso.}
\newblock \emph{The Annals of Statistics}, \textbf{34}(3), 1436--1462.

\bibitem[{Nicholson \emph{et~al.}(2014)Nicholson, Cozman, Velikova, van
  Scheltinga, Lucas, and Spaanderman}]{velikova2014}
Nicholson A, Cozman F, Velikova M, van Scheltinga JT, Lucas PJ, Spaanderman M
  (2014).
\newblock \enquote{Applications of Bayesian Networks Exploiting causal
  functional relationships in Bayesian Network Modelling for Personalised
  Healthcare.}
\newblock \emph{International Journal of Approximate Reasoning},
  \textbf{55}(1), 59--73.

\bibitem[{Niinim{\"a}ki \emph{et~al.}(2016)Niinim{\"a}ki, Parviainen, and
  Koivisto}]{niinimaki2016}
Niinim{\"a}ki T, Parviainen P, Koivisto M (2016).
\newblock \enquote{Structure Discovery in Bayesian Networks by Sampling Partial
  Orders.}
\newblock \emph{Journal of Machine Learning Research}, \textbf{17}(1),
  2002--2048.

\bibitem[{Pearl(2000)}]{Pearl00}
Pearl J (2000).
\newblock \emph{Causality: Models, Reasoning, and Inference}.
\newblock Cambridge University Press.

\bibitem[{Pe{\'e}r \emph{et~al.}(2001)Pe{\'e}r, Regev, Elidan, and
  Friedman}]{pe2001inferring}
Pe{\'e}r D, Regev A, Elidan G, Friedman N (2001).
\newblock \enquote{Inferring Subnetworks from Perturbed Expression Profiles.}
\newblock \emph{Bioinformatics}, \textbf{17}, S215--S224.

\bibitem[{Perrier \emph{et~al.}(2008)Perrier, Imoto, and Miyano}]{perrier2008}
Perrier E, Imoto S, Miyano S (2008).
\newblock \enquote{Finding Optimal Bayesian Network Given a Super-Structure.}
\newblock \emph{Journal of Machine Learning Research}, \textbf{9}(Oct),
  2251--2286.

\bibitem[{Pournara and Wernisch(2004)}]{pournara2004reconstruction}
Pournara I, Wernisch L (2004).
\newblock \enquote{Reconstruction of Gene Networks Using Bayesian Learning and
  Manipulation Experiments.}
\newblock \emph{Bioinformatics}, \textbf{20}(17), 2934--2942.

\bibitem[{{R Core Team}(2016)}]{rcore2016}
{R Core Team} (2016).
\newblock \emph{\proglang{R}: A Language and Environment for Statistical
  Computing}.
\newblock R Foundation for Statistical Computing, Vienna, Austria.
\newblock \urlprefix\url{https://www.R-project.org/}.

\bibitem[{Ravikumar \emph{et~al.}(2010)Ravikumar, Wainwright, and
  Lafferty}]{ravikumar2010}
Ravikumar P, Wainwright MJ, Lafferty JD (2010).
\newblock \enquote{High-Dimensional Ising Model Selection Using
  $\ell_1$-Regularized Logistic Regression.}
\newblock \emph{The Annals of Statistics}, \textbf{38}(3), 1287--1319.

\bibitem[{Rosseel(2012)}]{rosseel2012}
Rosseel Y (2012).
\newblock \enquote{\pkg{lavaan}: An \proglang{R} Package for Structural
  Equation Modeling.}
\newblock \emph{Journal of Statistical Software}, \textbf{48}(1), 1--36.
\newblock ISSN 1548-7660.
\newblock \doi{10.18637/jss.v048.i02}.
\newblock
  \urlprefix\url{https://www.jstatsoft.org/index.php/jss/article/view/v048i02}.

\bibitem[{Russell and Norvig(1995)}]{russell1995artificial}
Russell S, Norvig P (1995).
\newblock \enquote{Artificial Intelligence: A Modern Approach.}

\bibitem[{R{\"u}timann and B{\"u}hlmann(2009)}]{rutimann2009}
R{\"u}timann P, B{\"u}hlmann P (2009).
\newblock \enquote{High Dimensional Sparse Covariance Estimation via Directed
  Acyclic Graphs.}
\newblock \emph{Electronic Journal of Statistics}, \textbf{3}, 1133--1160.

\bibitem[{Sachs \emph{et~al.}(2005)Sachs, Perez, Pe'er, Lauffenburger, and
  Nolan}]{sachs2005}
Sachs K, Perez O, Pe'er D, Lauffenburger DA, Nolan GP (2005).
\newblock \enquote{Causal Protein-Signaling Networks Derived from
  Multiparameter Single-Cell Data.}
\newblock \emph{Science}, \textbf{308}(5721), 523--529.

\bibitem[{Sanford and Moosa(2012)}]{sanford2012}
Sanford AD, Moosa IA (2012).
\newblock \enquote{A Bayesian Network Structure for Operational Risk Modelling
  in Structured Finance Operations.}
\newblock \emph{Journal of the Operational Research Society}, \textbf{63}(4),
  431--444.

\bibitem[{Schmidt \emph{et~al.}(2007)Schmidt, Niculescu-Mizil, and
  Murphy}]{schmidt2007}
Schmidt M, Niculescu-Mizil A, Murphy K (2007).
\newblock \enquote{Learning Graphical Model Structure Using {L1}-Regularization
  Paths.}
\newblock In \emph{AAAI}, volume~7, pp. 1278--1283.

\bibitem[{Scutari(2010)}]{scutari2010}
Scutari M (2010).
\newblock \enquote{Learning {B}ayesian Networks with the \pkg{bnlearn}
  \proglang{R} Package.}
\newblock \emph{Journal of Statistical Software}, \textbf{35}(i03).
\newblock \urlprefix\url{https://www.jstatsoft.org/article/view/v035i03}.

\bibitem[{Shannon \emph{et~al.}(2003)Shannon, Markiel, Ozier, Baliga, Wang,
  Ramage, Amin, Schwikowski, and Ideker}]{shannon2003}
Shannon P, Markiel A, Ozier O, Baliga NS, Wang JT, Ramage D, Amin N,
  Schwikowski B, Ideker T (2003).
\newblock \enquote{Cytoscape: A Software Environment for Integrated Models of
  Biomolecular Interaction Networks.}
\newblock \emph{Genome research}, \textbf{13}(11), 2498--2504.

\bibitem[{Shannon \emph{et~al.}(2013)Shannon, Grimes, Kutlu, Bot, and
  Galas}]{shannon2013}
Shannon PT, Grimes M, Kutlu B, Bot JJ, Galas DJ (2013).
\newblock \enquote{RCytoscape: Tools for Exploratory Network Analysis.}
\newblock \emph{BMC bioinformatics}, \textbf{14}(1), 217.

\bibitem[{Spirtes and Glymour(1991)}]{spirtes1991}
Spirtes P, Glymour C (1991).
\newblock \enquote{An Algorithm for Fast Recovery of Sparse Causal Graphs.}
\newblock \emph{Social Science Computer Review}, \textbf{9}(1), 62--72.

\bibitem[{Spirtes \emph{et~al.}(2000)Spirtes, Glymour, and
  Scheines}]{spirtes2000}
Spirtes P, Glymour C, Scheines R (2000).
\newblock \emph{Causation, Prediction, and Search}, volume~81.
\newblock The MIT Press.

\bibitem[{Suzuki(1993)}]{Suzuki93}
Suzuki J (1993).
\newblock \enquote{A Construction of {B}ayesian Networks from Databases Based
  on an {MDL} Principle.}
\newblock In \emph{Proceedings of the Ninth Annual Conference on Uncertainty in
  Artificial Intelligence}, pp. 266--273.

\bibitem[{Tibshirani(1996)}]{tibshirani1996}
Tibshirani R (1996).
\newblock \enquote{Regression Shrinkage and Selection via the Lasso.}
\newblock \emph{Journal of the Royal Statistical Society B}, pp. 267--288.

\bibitem[{Tsamardinos \emph{et~al.}(2006)Tsamardinos, Brown, and
  Aliferis}]{tsamardinos2006max}
Tsamardinos I, Brown LE, Aliferis CF (2006).
\newblock \enquote{The Max-Min Hill-Climbing Bayesian Network Structure
  Learning Algorithm.}
\newblock \emph{Machine learning}, \textbf{65}(1), 31--78.

\bibitem[{Uhler \emph{et~al.}(2013)Uhler, Raskutti, B{\"u}hlmann, and
  Yu}]{uhler2013}
Uhler C, Raskutti G, B{\"u}hlmann P, Yu B (2013).
\newblock \enquote{Geometry of the Faithfulness Assumption in Causal
  Inference.}
\newblock \emph{The Annals of Statistics}, \textbf{41}(2), 436--463.

\bibitem[{van~de Geer and B{\"u}hlmann(2013)}]{geer2013}
van~de Geer S, B{\"u}hlmann P (2013).
\newblock \enquote{$\ell_0$-Penalized Maximum Likelihood for Sparse Directed
  Acyclic Graphs.}
\newblock \emph{The Annals of Statistics}, \textbf{41}(2), 536--567.

\bibitem[{Venables and Ripley(2002)}]{venables2002}
Venables WN, Ripley BD (2002).
\newblock \emph{Modern Applied Statistics with S}.
\newblock Fourth edition. Springer-Verlag, New York.
\newblock ISBN 0-387-95457-0,
  \urlprefix\url{http://www.stats.ox.ac.uk/pub/MASS4}.

\bibitem[{Wu and Lange(2008)}]{wu2008}
Wu TT, Lange K (2008).
\newblock \enquote{Coordinate descent algorithms for {L}asso penalized
  regression.}
\newblock \emph{The Annals of Applied Statistics}, pp. 224--244.

\bibitem[{Xiang and Kim(2013)}]{xiang2013}
Xiang J, Kim S (2013).
\newblock \enquote{A* {L}asso for Learning a Sparse {B}ayesian Network
  Structure for Continuous Variables.}
\newblock In \emph{Advances in Neural Information Processing Systems}, pp.
  2418--2426.

\bibitem[{Yang \emph{et~al.}(2015)Yang, Ravikumar, Allen, and Liu}]{yang2015}
Yang E, Ravikumar P, Allen GI, Liu Z (2015).
\newblock \enquote{Graphical Models via Univariate Exponential Family
  Distributions.}
\newblock \emph{Journal of Machine Learning Research}, \textbf{16}, 3813--3847.

\bibitem[{Yuan and Lin(2006)}]{yuan2006}
Yuan M, Lin Y (2006).
\newblock \enquote{Model Selection and Estimation in Regression with Grouped
  Variables.}
\newblock \emph{Journal of the Royal Statistical Society B}, \textbf{68}(1),
  49--67.

\bibitem[{Zhang \emph{et~al.}(2013)Zhang, Gaiteri, Bodea, Wang, McElwee,
  Podtelezhnikov, Zhang, Xie, Tran, Dobrin \emph{et~al.}}]{zhang2013}
Zhang B, Gaiteri C, Bodea LG, Wang Z, McElwee J, Podtelezhnikov AA, Zhang C,
  Xie T, Tran L, Dobrin R, \emph{et~al.} (2013).
\newblock \enquote{Integrated Systems Approach Identifies Genetic Nodes and
  Networks in Late-Onset Alzheimer's Disease.}
\newblock \emph{Cell}, \textbf{153}(3), 707--720.

\bibitem[{Zhang(2010)}]{zhang2010}
Zhang CH (2010).
\newblock \enquote{Nearly Unbiased Variable Selection Under Minimax Concave
  Penalty.}
\newblock \emph{The Annals of Statistics}, \textbf{38}(2), 894--942.

\bibitem[{Zhang(2016)}]{zhang2016}
Zhang D (2016).
\newblock \emph{Concave Penalized Estimation of Causal Gaussian Networks with
  Intervention}.
\newblock Master's thesis, UCLA. Statistics 0891.

\bibitem[{Zhang and Spirtes(2002)}]{zhang2002}
Zhang J, Spirtes P (2002).
\newblock \enquote{Strong Faithfulness and Uniform Consistency in Causal
  Inference.}
\newblock In \emph{Proceedings of the nineteenth conference on uncertainty in
  artificial intelligence}, pp. 632--639.

\bibitem[{Zhou(2011)}]{Zhou11}
Zhou Q (2011).
\newblock \enquote{Multi-Domain Sampling with Applications to Structural
  Inference of {B}ayesian Networks.}
\newblock \emph{Journal of the American Statistical Association}, \textbf{106},
  1317--1330.

\end{thebibliography}

\end{document}